%% file: Main.tex
\documentclass[twocolumn, switch]{article}
\usepackage{preprint}
\usepackage{algorithm}
\usepackage{algorithmic}
\usepackage{amsmath, amsthm, amssymb, amsfonts}

\usepackage[numbers,square]{natbib}
\usepackage[utf8]{inputenc}
\usepackage[T1]{fontenc}
\usepackage{xcolor}
\usepackage[colorlinks = true,
            linkcolor = purple,
            urlcolor  = blue,
            citecolor = cyan,
            anchorcolor = black]{hyperref}
\usepackage{booktabs}
\usepackage{nicefrac}
\usepackage{microtype}
\usepackage{lineno}
\usepackage{float}
\usepackage{lipsum}
\usepackage{newfloat}
\DeclareFloatingEnvironment[name={Supplementary Figure}]{suppfigure}
\usepackage{sidecap}
\sidecaptionvpos{figure}{c}

\usepackage{titlesec}
\titlespacing\section{0pt}{12pt plus 3pt minus 3pt}{1pt plus 1pt minus 1pt}
\titlespacing\subsection{0pt}{10pt plus 3pt minus 3pt}{1pt plus 1pt minus 1pt}
\titlespacing\subsubsection{0pt}{8pt plus 3pt minus 3pt}{1pt plus 1pt minus 1pt}

\usepackage{eso-pic}
\usepackage{tikz}
\usepackage{xcolor}
\PassOptionsToPackage{colorlinks=true,linkcolor=gray,urlcolor=gray}{hyperref}

\usepackage{titling}
\usepackage{footmisc}
\usepackage{multirow}
\usepackage{arydshln,mathtools,enumitem}
\setlist[enumerate,1]{label=\textbf{A\arabic*:}, leftmargin=*}
\usepackage{amsmath}

\newtheoremstyle{BoldRemark}{\topsep}{\topsep}{\normalfont}{}{\bfseries}{}{ }{}
\theoremstyle{BoldRemark}

\newcommand{\Author}[2]{\textbf{#1}\textsuperscript{#2}}

\title{BatMIL: Geometry-Embedding State Space Model for Whole-Slide Image Representation}

\author{
  \Author{Enhui Chai}{1} \and
  \Author{Sicheng Chen}{2}\and
  \Author{Tianyi Zhang}{3}\and
  \Author{Chad Wong}{2} \and
  \Author{Kecheng Huang}{1} \and
  \Author{Zeyu Liu}{1}\and
  \Author{Fei Xia}{2}
}

\date{
  \textsuperscript{1}PuzzleLogic Pte Ltd, Singapore 229594, Singapore \\
  \textsuperscript{2}Nhu Department of Electrical Engineering and Computer Science, University of California, Irvine \\
  \textsuperscript{3}Department of Electrical \& Computer Engineering, National University of Singapore \\
  [1em]
  \footnotesize Enhui Chai and Sicheng Chen contributed equally to this work. \\
  \footnotesize \textbf{Corresponding author:} Zeyu Liu\texttt{<zeyuliu@puzzlelogic.com>}, Fei Xia\texttt{<fei.xia@uci.edu>} \\
}

\begin{document}

\twocolumn[\begin{@twocolumnfalse}
\maketitle
\thispagestyle{empty}
\input{Sections/0_Abstract}
\vspace{0.35cm}
\end{@twocolumnfalse}]
\input{Sections/1_Introduction}
\input{Sections/2_Related_Work}
\input{Sections/3_Methods}
\input{Sections/4_Experiments}

\input{Sections/5_Discussion}

\input{Sections/6_Conclusion}
\bibliography{Sections/References}

\end{document}

%% file: Sections/0_Abstract.tex
\begin{abstract}
Accurate analysis of histopathological images is critical for disease diagnosis and treatment planning. Whole-slide images (WSIs), which digitize tissue specimens at gigapixel resolution, are fundamental to this process but require aggregating thousands of patches for slide-level predictions. Multiple Instance Learning (MIL) tackles this challenge with a two-stage paradigm, decoupling tile-level embedding and slide-level prediction. However, most existing methods implicitly embed patch representations in homogeneous Euclidean spaces, overlooking the hierarchical organization and regional heterogeneity of pathological tissues. This limits current models' ability to capture global tissue architecture and fine-grained cellular morphology. To address this limitation, we introduce a hybrid hyperbolic-Euclidean representation that embeds WSI features in dual geometric spaces, enabling complementary modeling of hierarchical tissue structures and local morphological details. Building on this formulation, we develop BatMIL, an MIL method that leverages both geometric spaces. To model long-range dependencies among thousands of patches, we employ a structured state space sequence model backbone that encodes patch sequences with linear computational complexity. Furthermore, to account for regional heterogeneity, we introduce a mixture-of-experts module that dynamically routes each patch to specialized expert sub-networks based on its features, improving representational capacity while reducing redundant computation. Experiments on seven datasets spanning six cancer types demonstrate that BatMIL achieves outstanding performance against fifteen state-of-the-art MIL approaches, demonstrating that geometry-embedding representation learning offers a promising direction for computational pathology.
\end{abstract}

\keywords{Computational pathology, whole-slide images, multiple instance learning, hybrid geometric embedding}

%% file: Sections/1_Introduction.tex
\section{Introduction}
Histopathological examination is the gold standard for disease diagnosis, forming the foundation for accurate cancer diagnosis and treatment planning~\cite{wsi}. Whole-slide images (WSIs) are central to computational pathology, yet their gigapixel resolution (e.g., $30,000 \times 50,000$ pixels) poses significant computational challenges. To tackle this gigapixel challenge, Multiple Instance Learning (MIL)~\cite{MIL} has emerged as a widely adopted solution by decoupling tile-level feature extraction from slide-level aggregation. It partitions WSIs into small regions (e.g., $224 \times 224$ pixels), generates patch embeddings with foundation models (e.g., UNI~\cite{chen2024uni}, Virchow2~\cite{zimmermann2024virchow2}), and aggregates them using slide-level models (e.g., ABMIL~\cite{abmil}, TransMIL~\cite{transmil}). Although practical, existing MIL methods primarily address computational scalability, while the intrinsic biological properties of WSIs remain insufficiently modeled. Specifically, they suffer from the following three key limitations:

\begin{figure}[t]
    \centering
    \includegraphics[width=\linewidth]{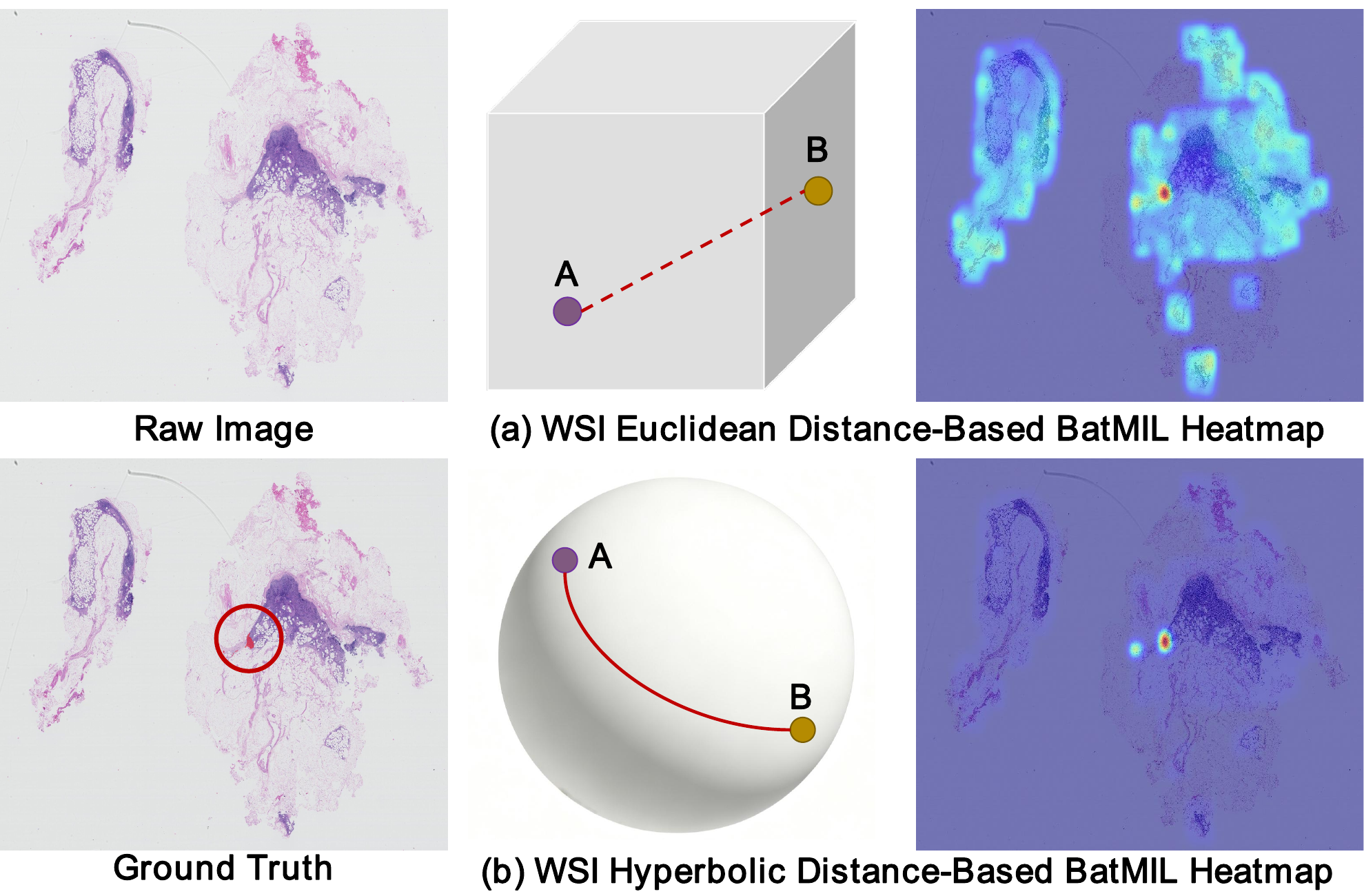}
    \caption{WSI geometric representations. \textbf{(a)} Euclidean distance fails to model hierarchical relationships. \textbf{(b)} Hyperbolic geometry naturally encodes the hierarchical structure of pathology.}
    \label{fig:differ}
\end{figure}

\begin{figure*}[t]
    \centering
    \includegraphics[width=0.75\textwidth]{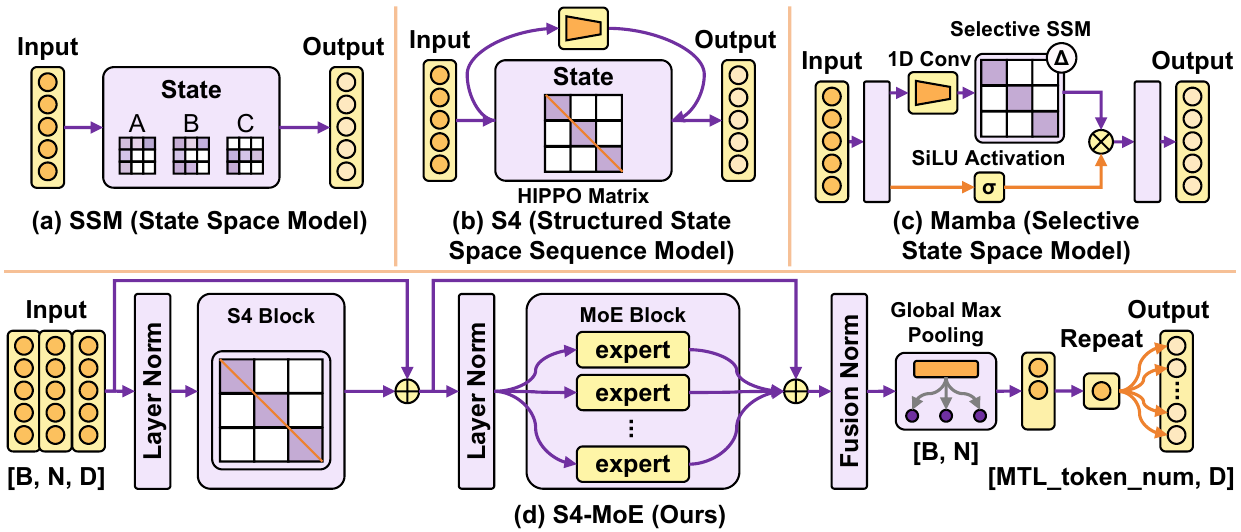}
    \caption{Comparison of SSM structures. \textbf{(a)} SSM (State Space Model), \textbf{(b)} S4 (Structured State Space Sequence Model), \textbf{(c)} Mamba (Selective State Space Model), \textbf{(d)} S4-MoE (Ours)}.
    \label{fig:S4-MoE}
\end{figure*}

\noindent \textit{Geometric trade-offs in hierarchy-sensitive slide-level representation.} 
Beyond spatial dependencies and heterogeneity, pathological tissues inherently exhibit a multi-level hierarchical architecture (e.g., cells $\rightarrow$ glands $\rightarrow$ tissues). Most traditional MIL methods rely on Euclidean distance, which excels at capturing locally continuous morphological features but is not designed to respect the exponential growth patterns characteristic of such hierarchies. Consequently, Euclidean-based activation maps often scatter attention across diagnostic-irrelevant background or stromal regions (Fig.~\ref{fig:differ} (a)). To address this, recent pioneering works~\cite{huang2025hyperpath} have introduced hyperbolic geometry to WSI analysis, exploiting its inherent capacity to embed hierarchical topologies. However, pure hyperbolic representations face a fundamental geometric trade-off: their inherent negative curvature distorts fine-grained local morphological details that are naturally better preserved in flat Euclidean spaces. This motivates a hybrid geometric representation that seeks to balance these two objectives, precisely concentrating heatmap activations on tumor boundaries while suppressing irrelevant background noise (Fig.~\ref{fig:differ} (b)).

\noindent \textit{Inadequate modeling of long-range 2D spatial dependencies.} 
Effectively capturing global tissue structures requires modeling inter-tile relationships across thousands of patches, which State Space Models (SSMs) excel at (Fig.~\ref{fig:S4-MoE} (a)). Among them, the Structured State Space Sequence (S4) model~\cite{s4} is an early and representative formulation that offers stable long-context modeling through structured state-space parameterization. Later selective SSMs such as Mamba further improve sequence modeling efficiency and adaptivity, but their selective scanning mechanism still imposes an ordered 1D traversal over spatial tokens, which can introduce directional bias and structurally conflicts with the non-directional 2D grid topology of WSIs (Fig.~\ref{fig:S4-MoE} (c)). In contrast, S4 provides a more suitable basis for our setting: its structured global recurrence can model long-range interactions without relying on rigid scan directions, thereby better aligning with the complex 2D spatial organization of pathological tissues while maintaining linear complexity and stable global receptive fields (Fig.~\ref{fig:S4-MoE} (b)).

\noindent \textit{Vulnerability to pronounced regional heterogeneity.}
A single WSI is a highly heterogeneous microenvironment composed of morphologically distinct regions, such as dense tumor areas, necrotic zones, and normal stroma. Traditional MIL methods attempt to fit these divergent distributions using a single unified network backbone. This often leads to parameter dilution and task interference, resulting in strong performance in dominant regions but severe bias or underfitting in others~\cite{zhang2022dtfd}. Effectively disentangling these highly varied histological features remains a significant bottleneck, necessitating a dynamic routing mechanism that can adaptively assign specialized computational resources to distinct morphological patterns.

Building upon these insights, we propose \textbf{BatMIL} (Fig.~\ref{fig:main}), a novel MIL framework that conceptually bridges S4, MoE, and geometric representation learning to systematically address the aforementioned limitations. Specifically, we design a cascaded backbone network that utilizes S4 as the core for unbiased 2D global context modeling, while introducing a mixture-of-experts (MoE) module to dynamically route heterogeneous regional features. Furthermore, we develop a hybrid geometric representation module to jointly embed WSI features into a hyperbolic-Euclidean space, fully encapsulating the hierarchical tissue topology without sacrificing local details. Our main contributions are summarized as follows:

\begin{itemize}
    \item We propose BatMIL, a geometry-embedding MIL framework that addresses the challenges of ultra-long spatial modeling, regional heterogeneity, and hierarchical tissue topologies in WSI analysis.
    \item We develop a cascaded S4-MoE backbone (Fig.~\ref{fig:S4-MoE} (d)). The S4 module acts as an unbiased engine for capturing global 2D context with linear complexity, while the MoE module dynamically routes heterogeneous tissue patches to specialized experts, mitigating feature homogenization.
    \item We design a hybrid hyperbolic-Euclidean representation module. By harmonizing the hierarchical embedding capacity of hyperbolic geometry with the local morphological sensitivity of Euclidean space, this dual-space approach significantly concentrates diagnostic attention and suppresses irrelevant background noise.
    \item Experiments on seven WSI datasets spanning six cancer types demonstrate that BatMIL achieves strong performance against fifteen MIL methods, supporting the effectiveness of structured geometric representation for computational pathology.
\end{itemize}

%% file: Sections/2_Related_Work.tex
\section{Related Work}
\subsection{Multiple Instance Learning}
Deep learning for WSI analysis has evolved toward a two-stage MIL paradigm, separating tile-level embedding and slide-level modeling. ABMIL~\cite{abmil} addresses it by implementing a gated attention mechanism to adaptively weight informative tiles. Building on this, CLAM~\cite{clam} imposes instance-clustering constraints to encourage diverse and discriminative feature learning, thereby improving model generalization. Furthermore, DSMIL~\cite{dsmil} incorporates contrastive learning by combining instance- and bag-level supervision to better distinguish informative tiles. On the other hand, DTFD-MIL~\cite{dtfd} proposes a double-tier feature distillation framework that utilizes pseudo-bags to virtually expand the training set, enhancing robustness in small-sample scenarios.

However, traditional MIL methods often assume instances are independent, failing to capture the spatial and morphological inter-tile relationships. To address this, graph-based and multi-scale hierarchical methods were introduced. Patch-GCN~\cite{patchgcn} hierarchically aggregates instance-level histology features via a Graph Convolutional Network (GCN) to model local and global topological structures. Similarly, ZoomMIL~\cite{zoommil} builds WSI representations by aggregating tissue-context information from multiple magnifications, while HiGT~\cite{higt} introduces a hierarchical graph transformer to learn short-range local details and long-range global patterns. Despite successfully capturing spatial contexts, these methods rely heavily on explicitly defined graph topologies or rigid scales, limiting their flexibility.

To model global patch correlations without manual topological priors, Transformer-based methods have been widely adopted. TransMIL~\cite{transmil} overcomes the instance independence challenge by introducing a Transformer-based architecture, utilizing the Nyström approximation of the self-attention mechanism to explicitly model these correlations with pseudo-linear complexity. However, for the extreme sequence lengths inherent in gigapixel WSIs, such approximations still face memory bottlenecks and degradation in capturing fine-grained long-range dependencies. To explicitly improve long-context understanding, GigaPath~\cite{gigapath} adopts a cascaded structure and leverages the LongNet architecture. While its dilated attention effectively scales to extreme sequence lengths, the massive parameter count of GigaPath imposes a severe computational hardware burden compared to lightweight aggregators. 

Recently, State Space Models (SSMs) have emerged as a promising paradigm to break the computational bottlenecks of attention mechanisms while maintaining effective global receptive fields. Leveraging true linear complexity, S4MIL~\cite{s4mil} introduces the structured state space sequence (S4) model to WSI analysis, providing robust long-range dependency modeling. Subsequent works, such as MambaMIL~\cite{mambamil} and MamMIL~\cite{mammil}, adapt the Mamba architecture to capture inter-tile dependencies efficiently through selective scanning mechanisms, while PathRWKV~\cite{chen2025pathrwkv} explores linear recurrent neural networks for scalable WSI modeling. 

However, all of these previous methods solely embedded their feature representations in Euclidean space. This geometric constraint prevents them from explicitly capturing the intrinsic tree-like hierarchical topology of pathological structures. Furthermore, these unified network architectures lack the specialized capacity to dynamically adapt to the pronounced regional heterogeneity present in WSIs.

\subsection{Mixture of Experts in Transformers}
Mixture of Experts (MoE) provides an elegant paradigm for efficient model scaling by decoupling computational cost from total parameter count~\cite{liu2025stainexpert}. In Transformer architectures, this is achieved by replacing standard dense Feed-Forward Networks (FFNs) with a sparse ensemble of parallel "expert" subnetworks. For each input token, a lightweight gating network (router) dynamically activates only a sparse subset of these experts. Following its initial integration into Transformers via GShard~\cite{lepikhin2020gshard}, subsequent architectures like the Switch Transformer~\cite{fedus2022switch} streamlined the paradigm using a top-1 routing mechanism, coupled with auxiliary load-balancing losses to mitigate expert collapse. More recently, models such as Mixtral 8x7B~\cite{jiang2024mixtral} have demonstrated the robust scalability of top-k routing, achieving state-of-the-art performance. Beyond natural language processing, Vision MoE (V-MoE)~\cite{riquelme2021scaling} extended this architecture to computer vision by routing individual image patches. In the context of computational pathology, this dynamic patch-level routing offers a highly promising solution to the pronounced regional heterogeneity of WSIs, allowing morphologically distinct tissue patches to be adaptively processed by specialized experts rather than a constrained unified backbone.

\subsection{Geometric Representation Learning}
Recent advances in WSI analysis have increasingly recognized the limitations of conventional Euclidean representations, motivating the exploration of alternative representation spaces and feature transformation strategies. For instance, CI-MIL~\cite{cimil} introduces causal feature distillation and decorrelation to strengthen the relationship between slide-level predictions and diagnostically relevant regions. Meanwhile, hyperbolic representation learning has emerged as a promising direction for modeling the intrinsic hierarchical organization of pathological tissues. Methods such as HyperPath~\cite{huang2025hyperpath} and HVHM~\cite{HVHM} embed WSI features into hyperbolic spaces to exploit their capacity for representing tree-like tissue hierarchies with low distortion. More recently, H2-Surv~\cite{H2-Surv} further explores hierarchical hyperbolic representations in a multimodal setting for survival prediction. However, existing hyperbolic approaches predominantly rely on hyperbolic representations~\cite{ramasinghe2024accept,HVT}. Consequently, effectively harmonizing Euclidean and hyperbolic geometries to jointly model global hierarchical structures while preserving local feature discriminability remains a critical open challenge.

%% file: Sections/3_Methods.tex
\begin{figure*}[t]
    \centering
    
    \includegraphics[width=\linewidth]{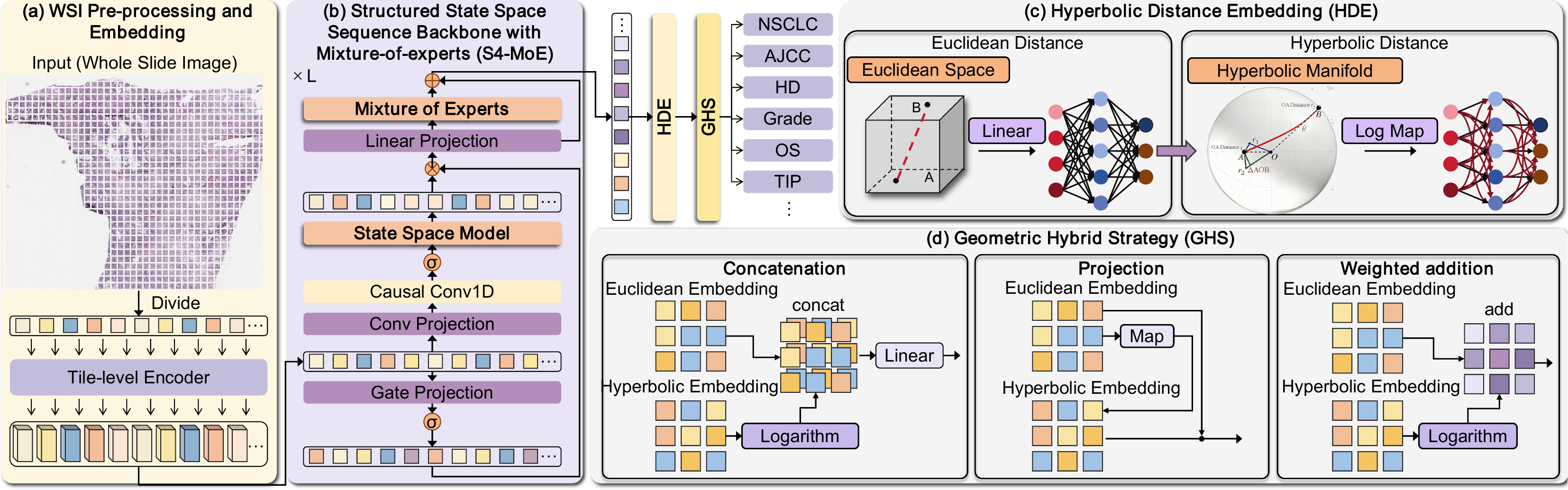}
    \caption{Overview of BatMIL. 
    \textbf{(a)} A WSI is first partitioned into tiles, then encoded into embeddings with a pathological foundation model. 
    \textbf{(b)} S4-MoE employs WSI-specialized experts to extract domain-specific features. 
    \textbf{(c)} The HDE module maps features from Euclidean space into a hyperbolic manifold to capture hierarchical relationships. 
    \textbf{(d)} The GHS module integrates embeddings from both hyperbolic and Euclidean geometries to generate a multi-scale representation.}
    \label{fig:main}
    
\end{figure*}

\section{Methods}
\subsection{WSI Pre-processing and Embedding}
To enable effective feature extraction and subsequent slide-level analysis on WSIs, we adopted a systematic preprocessing pipeline to filter uninformative regions and generate high-quality tiles. For a given WSI, we first load it at 0.5 microns-per-pixel (mpp) resolution. To avoid redundant computation on the background, we compute the luminance of the slide and apply Otsu's thresholding to generate a binary foreground mask. The slide is then cropped into bounding boxes encompassing the tissue regions.

As shown in Fig.~\ref{fig:main}(a), these bounding boxes are split into a grid of non-overlapping tiles, each with a size of $224 \times 224$. The WSI can then be represented as a set of tiles $\{P_{i,j}\}$. To ensure these samples are useful for pathological analysis, we adopt a rigorous screening process to exclude low-quality samples. First, we discard any tile where the tissue occupancy is below 10\%. Second, we eliminate tiles with insufficient pixel variation by computing the standard deviation across the RGB channels; tiles with an average standard deviation of less than 5 on a 0–255 scale are removed. Additionally, we filter out anomalous tiles containing excessive pure background, specifically discarding those where more than 50\% of the pixels have zero intensity.

The remaining tiles contain meaningful pathological information, and we subsequently utilize the tile-level encoder from GigaPath~\cite{gigapath} to convert them into compact semantic feature vectors. This feature extraction step improves both the training efficiency of the downstream slide-level backbone and its overall performance in WSI-level analysis.

\subsection{Structured State Space Sequence Module}\label{S4_B}
In BatMIL, we adopt the Structured State Space Sequence (S4) model~\cite{s4} as the backbone for long-range WSI modeling. S4 is a foundational State Space Model built upon a continuous-time linear state-space system. By discretizing the continuous-time matrices, S4 transforms the recurrence into a structured global convolution kernel, enabling parallel long-range modeling with linear complexity.
\begin{equation}
\dot{s}(t) = A s(t) + B x(t), \qquad
y(t) = C s(t) + D x(t),
\label{eq:ssm_continuous}
\end{equation}
where $x(t)$ and $y(t)$ denote the input and output, respectively, and $s(t)$ represents the latent state. After discretization, the recurrence can be written as
\begin{equation}
s_t = \bar{A}s_{t-1} + \bar{B}x_t, \qquad
y_t = Cs_t + Dx_t.
\label{eq:s4_recurrence}
\end{equation}
By exploiting a structured state matrix initialized from HiPPO (High-order Polynomial Projection Operators), S4 can efficiently model long-range dependencies. Importantly, the recurrent formulation can be equivalently expressed as a structured global convolution:
\begin{equation}
y = K * x, \qquad
K = \left(C\bar{B},\, C\bar{A}\bar{B},\, \ldots,\, C\bar{A}^{L-1}\bar{B}\right),
\label{eq:s4_conv}
\end{equation}
which enables efficient long-sequence modeling without the quadratic complexity of standard self-attention.

Although Mamba~\cite{mamba}, a recent selective SSM, has been widely adopted in MIL due to its efficient selective scan mechanism, we choose S4 because its inductive bias better matches both WSI characteristics and our geometric representation. Mamba employs input-dependent selective dynamics, making feature propagation more sensitive to the imposed scan order. In contrast, S4 uses shared state-space parameters and a stationary global kernel, which is more suitable for WSIs that lack an intrinsic causal ordering. This property is also conceptually consistent with our hyperbolic representation, where the relationship between two embeddings is symmetric and depends on their relative positions rather than traversal order. Therefore, S4 provides a more natural backbone for our geometry-embedding framework.

\subsection{MoE Routing}
WSIs are highly heterogeneous microenvironments comprising distinct regions. Traditional unified networks often suffer from parameter dilution when fitting these divergent distributions. To address this pronounced regional heterogeneity, the updated features from the S4 branch are fed into a sparse Mixture-of-Experts (MoE) module. Mixture-of-Experts~\cite{moe} is a neural architecture that increases model capacity without a proportional increase in computational cost, as it dynamically routes each input token to a sparse subset of parallel expert networks through a learnable gating function. Such sparse routing is particularly suitable for WSIs, where morphologically heterogeneous regions may benefit from specialized feature transformations.

We instantiate $k=4$ expert sub-networks, each formulated as a two-layer feedforward network with GELU activation and dropout to enhance non-linear expressiveness. Instead of routing all patches through the same dense layers, a lightweight gating network dynamically selects the top-2 specialized experts for each patch based on its contextualized representation from the S4 backbone. The gating network computes logits that are normalized via softmax to yield routing weights $w_j$. The outputs of the selected experts are then aggregated and residually fused. 
\begin{equation}
x_{MoE} = x_{S4} + \sum_{j \in I} w_j \cdot \text{E}_j\left(\text{LN}(x_{S4})\right)
\label{eq:moe}
\end{equation}
where $I$ is the set of selected expert indices, $w_j$ is the gating weight, and $\text{E}_j$ represents the $j$-th expert. Through this dynamic sparse activation, computational resources are adaptively assigned to morphologically distinct regions.

Following the cascaded S4-MoE backbone, we apply global max-pooling along the patch sequence dimension to aggregate the patch-level features into a unified slide-level representation, preserving the most discriminative spatial responses:
\begin{equation}
x_{\text{slide}} = \text{MaxPool}(x_{MoE}, \text{dim}=1)
\label{eq:pool}
\end{equation}
The slide-level feature $x_{\text{slide}}$ is then repeated to match the token count required for subsequent multi-task learning, yielding the final output tensor used for task-specific predictions.

\subsection{Hybrid Geometric Representation}
While the S4-MoE backbone effectively extracts slide-level representations, conventional MIL methods embed these features exclusively in Euclidean space. Euclidean geometry naturally preserves fine-grained local morphological details via the $L_2$ norm, but mathematically struggles to model the exponential tree-like hierarchical architecture intrinsic to pathological tissues. To capture this broader hierarchical scale, we introduce hyperbolic geometry via the Poincaré ball model $\mathbb{M}*c^n$, governed by the manifold curvature $c$. The Poincar'e ball is a standard model of hyperbolic geometry with constant negative curvature. Its distance metric $D*{\text{hyp}}$ is derived from Möbius addition ($\oplus_c$):
\begin{equation}
D_{\text{hyp}}(\mathbf{x}, \mathbf{y}) = \frac{2}{\sqrt{c}} \text{arctanh} \left( \sqrt{c}  \big\lVert -\mathbf{x} \oplus_c \mathbf{y} \big\rVert \right)
\label{eq:hyp_dist}
\end{equation}
where $\lVert \cdot \rVert_2$ denotes the Euclidean $L_2$ norm, with $\lVert\mathbf{x}\rVert_2=\sqrt{\sum_{i=1}^{n}x_i^2}$. This manifold embedding scales exponentially with the radius, effectively accommodating hierarchical tissue structures with minimal distortion. However, this global advantage introduces a geometric trade-off: pure hyperbolic space's inherent negative curvature can distort local discriminative features that are otherwise stable in flat Euclidean spaces.

Recognizing that hyperbolic geometry generalizes Euclidean geometry as the curvature approaches zero ($\lim_{c \to 0} D_{\text{hyp}}(\mathbf{x}, \mathbf{y}) = 2 \lVert\mathbf{x} - \mathbf{y}\rVert$), we propose a Geometric Hybrid Strategy (GHS) to harmonize both spaces. While fusion can technically be achieved via simple vector concatenation or complex distance projection mechanisms, we empirically adopt a weighted-addition strategy (Fig.~\ref{fig:main}d) for its efficiency and stability.

Given a slide-level feature $h$, we obtain Euclidean and hyperbolic embeddings as
\begin{equation}
z_E = W_E \cdot h, \qquad z_H = \exp_0^c(W_H \cdot h),
\label{eq:ghs_proj}
\end{equation}
where the exponential and logarithmic maps at the origin of the Poincar'e ball are defined as
\begin{equation}
\exp_0^c(\mathbf{v}) = \tanh(\sqrt{c} \cdot \lVert \mathbf{v} \rVert) \cdot \frac{\mathbf{v}}{\sqrt{c} \cdot \lVert \mathbf{v} \rVert},
\label{eq:exp_map}
\end{equation}
\begin{equation}
\log_0^c(\mathbf{y}) = \frac{1}{\sqrt{c}} \cdot \mathrm{arctanh}(\sqrt{c} \cdot \lVert \mathbf{y} \rVert) \cdot \frac{\mathbf{y}}{\lVert \mathbf{y} \rVert}.
\label{eq:log_map}
\end{equation}

Crucially, before the logarithmic map, the hyperbolic branch applies an explicit M\"obius linear transformation directly on the Poincar'e ball:
\begin{equation}
z_H' = \psi_c\left( \mathbf{W} \otimes_c z_H \oplus_c \mathbf{b} \right),
\label{eq:mobius_linear}
\end{equation}
where $\otimes_c$, $\oplus_c$, and $\psi_c$ denote M\"obius matrix-vector multiplication, M\"obius addition, and M\"obius pointwise nonlinearity, respectively.
\begin{equation}
\mathbf{W} \otimes_c z_H
=
\exp_0^c\left(
\mathbf{W}\cdot\log_0^c(z_H)
\right),
\label{eq:mobius_matvec}
\end{equation}
\begin{equation}
\begin{multlined}
\mathbf{x} \oplus_c \mathbf{y}
=
\big[
(1+2c\langle\mathbf{x},\mathbf{y}\rangle
+c\lVert\mathbf{y}\rVert^2)\mathbf{x}
\\
+(1-c\lVert\mathbf{x}\rVert^2)\mathbf{y}
\big]
\big/
\big[
1+2c\langle\mathbf{x},\mathbf{y}\rangle
\\
+c^2\lVert\mathbf{x}\rVert^2
\lVert\mathbf{y}\rVert^2
\big].
\end{multlined}
\label{eq:mobius_add}
\end{equation}
Here, $\psi_c$ applies the pointwise activation $\sigma(\cdot)$ (GELU in our implementation) in the tangent space and maps the result back to the Poincar'e ball.

Finally, the transformed hyperbolic feature is mapped back to the tangent space and fused with the Euclidean representation using a globally shared, feature-wise learnable gate $\alpha$ that is optimized during training and fixed at inference.
\begin{equation}
z_{\mathrm{hybrid}} = \alpha \cdot z_E + (1-\alpha) \cdot \log_0^c(z_H'),
\label{eq:ghs_fusion}
\end{equation}
where $\alpha=\mathrm{sigmoid}(\theta)\in(0,1)^m$, $\theta\in\mathbb{R}^m$, is a globally shared feature-wise learnable gate. It is jointly optimized during training and fixed at inference. This shared feature-wise fusion preserves the embedding dimensionality while balancing the hierarchical modeling capacity of the hyperbolic branch with the local morphological sensitivity of the Euclidean branch.

%% file: Sections/4_Experiments.tex
\section{Experiments}
\subsection{Datasets and Downstream Tasks}
We validate BatMIL on six core downstream tasks across seven WSI datasets, covering diverse diagnostic scenarios in computational pathology (Tab. \ref{tab:dataset}). In the \textbf{CAMELYON16}~\cite{cam16} (CAM16) and \textbf{CAMELYON17}~\cite{cam17} (CAM17) datasets, the \textbf{Breast Metastasis (BrMet)} task evaluates the model's capability to classify lymph nodes as normal or tumorous. The \textbf{PANDA}~\cite{panda} dataset is used for \textbf{ISUP grading (ISUP)} to assess prostate cancer aggressiveness, which requires fine-grained analysis of cellular patterns. The \textbf{TCGA-BLCA} dataset (BLCA) is employed to determine tumor differentiation levels in bladder cancer. The \textbf{TCGA-BRCA} dataset (BRCA) evaluates \textbf{IHC-HER2} status prediction task, a classification of HER2 expression levels that assesses suitability for targeted therapy. The \textbf{TCGA-CESC} dataset (CESC) is with the \textbf{Lymphovascular Invasion (LymInv)} task, which identifies the subtle presence of cancer invasion into lymphatic or vascular channels. The \textbf{TCGA-NSCLC} dataset is used to predict \textbf{AJCC-defined tumor staging (T-Stage)}, which reflects the tumor's local progression. Table~\ref{tab:dataset} reports 3,210 WSIs and 10 classes for this task.

\begin{table}[htbp]
\centering
\caption{Implemented datasets and corresponding downstream tasks.}
\label{tab:dataset}
\renewcommand{\arraystretch}{1.2}
\resizebox{\columnwidth}{!}{
\begin{tabular}{ccccc}
\hline
\textbf{Dataset} & \textbf{\# WSIs} & \textbf{Task} & \textbf{Organism} & \textbf{\# Classes} \\
\hline
\textbf{CAM16} & 400   & BrMet    & Lymph Node & 2  \\
\textbf{CAM17} & 500   & BrMet    & Lymph Node & 4  \\
\textbf{PANDA} & 10616 & ISUP  & Prostate   &  6  \\
\textbf{BLCA} & 926   & GRADE    & Bladder    & 2  \\
\textbf{BRCA} & 3121  & IHC-HER2 & Breast     & 4  \\
\textbf{CESC} & 604   & LymInv   & Cervix     & 2  \\
\textbf{NSCLC} & 3210  & T-Stage  & Lung       & 10 \\
\hline
\end{tabular}}

\end{table}

\begin{table*}[p]
\caption{Performance comparison with fifteen methods. (Note: Standard deviations $\sigma$ are represented by symbols: $^{\ddagger}$ for $\sigma \le 0.005$, $^{\dagger}$ for $0.005 < \sigma \le 0.015$, and $^*$ for $\sigma > 0.015$. The best and second-best results are highlighted in \textbf{bold} and \underline{underlined}.)}
\label{tab:comparison}
\centering
\setlength{\tabcolsep}{1pt}
\renewcommand{\arraystretch}{1.2}
\resizebox{\textwidth}{!}{
\begin{tabular}{c|c|cccc|ccc|cc|cccc|cc|c}
\hline
\multirow{2}{*}{\textbf{Dataset}} & \multirow{2}{*}{\textbf{Metric}} & \textbf{ABMIL} & \textbf{CLAM} & \textbf{DSMIL} & \textbf{DTFD-MIL} & \textbf{PatchGCN} & \textbf{ZoomMIL} & \textbf{HiGT} & \textbf{TransMIL} & \textbf{GigaPath} & \textbf{S4MIL} & \textbf{MambaMIL} & \textbf{MamMIL} & \textbf{PathRWKV} & \textbf{CI-MIL} & \textbf{H2-Surv} & \textbf{BatMIL} \\
&& \cite{abmil} & \cite{clam} & \cite{dsmil} & \cite{dtfd} & \cite{patchgcn} & \cite{zoommil} & \cite{higt} & \cite{transmil} & \cite{gigapath} & \cite{s4mil} & \cite{mambamil} & \cite{mammil} & \cite{chen2025pathrwkv} & \cite{cimil} & \cite{H2-Surv} & \textbf{(Ours)} \\
\hline
\multicolumn{18}{c}{\textbf{GigaPath}~\cite{gigapath}} \\
\hline
& \textbf{AUROC} & 0.989$^{\ddagger}$ & 0.990$^{\ddagger}$ & 0.965$^{\ddagger}$ & 0.990$^{\dagger}$ & 0.967$^{\ddagger}$ & 0.973$^{\ddagger}$ & 0.965$^{\ddagger}$ & \underline{0.993}$^{\ddagger}$ & 0.987$^{\dagger}$ & 0.989$^*$ & 0.992$^{\ddagger}$ & 0.986$^{\ddagger}$ & 0.991$^{\ddagger}$ & 0.988$^{\dagger}$ & 0.868$^{\ddagger}$ & \textbf{0.996}$^{\ddagger}$ \\
\textbf{CAM16} & \textbf{Accuracy} & 0.980$^{\ddagger}$ & 0.981$^{\ddagger}$ & 0.922$^{\ddagger}$ & 0.981$^{\ddagger}$ & 0.947$^{\ddagger}$ & 0.951$^{\ddagger}$ & 0.945$^{\dagger}$ & 0.982$^{\ddagger}$ & 0.961$^*$ & 0.984$^{\ddagger}$ & 0.983$^{\dagger}$ & 0.976$^{\ddagger}$ & 0.983$^{\dagger}$ & \underline{0.985}$^{\ddagger}$ & 0.798$^*$ & \textbf{0.988}$^{\ddagger}$ \\
& \textbf{F1-Score} & 0.978$^{\ddagger}$ & 0.980$^{\ddagger}$ & 0.915$^{\dagger}$ & 0.980$^{\ddagger}$ & 0.939$^{\ddagger}$ & 0.944$^{\ddagger}$ & 0.937$^{\ddagger}$ & 0.981$^{\ddagger}$ & 0.958$^*$ & \textbf{0.983}$^{\dagger}$ & \underline{0.982}$^{\dagger}$ & 0.975$^{\ddagger}$ & 0.981$^{\dagger}$ & \textbf{0.983}$^{\dagger}$ & 0.786$^{\ddagger}$ & 0.977$^*$ \\
\hline
& \textbf{AUROC} & 0.708$^*$ & 0.710$^*$ & 0.710$^*$ & 0.716$^*$ & 0.548$^{\dagger}$ & 0.587$^*$ & 0.577$^{\dagger}$ & 0.700$^*$ & 0.697$^*$ & 0.541$^*$ & 0.719$^*$ & 0.546$^{\dagger}$ & 0.722$^*$ & \underline{0.728}$^{\ddagger}$ & 0.550$^{\dagger}$ & \textbf{0.778}$^*$ \\
\textbf{CAM17} & \textbf{Accuracy} & 0.756$^*$ & 0.758$^*$ & 0.746$^*$ & 0.756$^{\dagger}$ & 0.545$^*$ & 0.583$^*$ & 0.570$^*$ & 0.754$^*$ & 0.731$^*$ & 0.646$^{\dagger}$ & 0.758$^*$ & 0.667$^*$ & 0.752$^*$ & \underline{0.768}$^{\ddagger}$ & 0.626$^{\dagger}$ & \textbf{0.778}$^*$ \\
& \textbf{F1-Score} & 0.498$^{\dagger}$ & \underline{0.503}$^*$ & 0.436$^*$ & 0.463$^*$ & 0.242$^*$ & 0.309$^*$ & 0.311$^*$ & 0.404$^*$ & 0.423$^*$ & 0.244$^*$ & 0.460$^*$ & 0.303$^*$ & 0.499$^*$ & 0.470$^{\ddagger}$ & 0.489$^{\ddagger}$ & \textbf{0.507}$^*$ \\
\hline
& \textbf{AUROC} & 0.946$^{\ddagger}$ & 0.947$^{\ddagger}$ & 0.942$^{\dagger}$ & 0.944$^{\ddagger}$ & 0.942$^{\ddagger}$ & 0.944$^{\dagger}$ & 0.940$^{\ddagger}$ & 0.936$^{\ddagger}$ & 0.940$^{\ddagger}$ & 0.940$^{\ddagger}$ & 0.942$^{\ddagger}$ & 0.948$^{\ddagger}$ & 0.947$^{\ddagger}$ & \underline{0.949}$^{\ddagger}$ & 0.939$^{\ddagger}$ & \textbf{0.960}$^{\ddagger}$ \\
\textbf{PANDA} & \textbf{Accuracy} & 0.768$^{\ddagger}$ & 0.774$^{\ddagger}$ & 0.763$^*$ & 0.762$^{\dagger}$ & 0.744$^{\dagger}$ & 0.749$^{\ddagger}$ & 0.741$^{\ddagger}$ & 0.739$^{\dagger}$ & 0.752$^{\dagger}$ & 0.759$^{\ddagger}$ & 0.759$^{\dagger}$ & 0.757$^{\dagger}$ & \underline{0.776}$^{\ddagger}$ & 0.772$^{\ddagger}$ & 0.723$^{\ddagger}$ & \textbf{0.798}$^{\ddagger}$ \\
& \textbf{F1-Score} & 0.711$^{\dagger}$ & 0.721$^{\dagger}$ & 0.701$^*$ & 0.705$^{\ddagger}$ & 0.705$^{\ddagger}$ & 0.711$^{\dagger}$ & 0.705$^{\dagger}$ & 0.672$^{\dagger}$ & 0.692$^{\dagger}$ & 0.707$^{\ddagger}$ & 0.693$^{\dagger}$ & 0.699$^{\ddagger}$ & \underline{0.726}$^{\dagger}$ & 0.717$^{\ddagger}$ & 0.666$^{\ddagger}$ & \textbf{0.748}$^{\dagger}$ \\
\hline
& \textbf{AUROC} & 0.983$^{\dagger}$ & 0.986$^{\ddagger}$ & 0.839$^{\dagger}$ & \underline{0.994}$^{\dagger}$ & 0.918$^{\dagger}$ & 0.922$^{\ddagger}$ & 0.913$^{\ddagger}$ & 0.985$^{\dagger}$ & 0.977$^{\dagger}$ & 0.942$^{\dagger}$ & 0.904$^{\ddagger}$ & 0.938$^{\dagger}$ & 0.991$^{\ddagger}$ & \underline{0.994}$^{\dagger}$ & 0.993$^{\dagger}$ & \textbf{0.997}$^{\ddagger}$ \\
\textbf{BLCA} & \textbf{Accuracy} & 0.932$^{\ddagger}$ & 0.943$^{\dagger}$ & 0.909$^{\ddagger}$ & 0.954$^{\dagger}$ & \underline{0.966}$^{\ddagger}$ & \textbf{0.977}$^{\ddagger}$ & 0.955$^{\ddagger}$ & 0.943$^{\dagger}$ & 0.943$^{\dagger}$ & 0.921$^{\dagger}$ & 0.943$^{\dagger}$ & 0.911$^{\dagger}$ & \textbf{0.977}$^{\ddagger}$ & 0.943$^{\dagger}$ &  0.932$^*$ & \textbf{0.977}$^{\dagger}$ \\
& \textbf{F1-Score} & 0.732$^*$ & 0.792$^*$ & 0.576$^*$ & \underline{0.845}$^{\dagger}$ & 0.667$^{\dagger}$ & 0.731$^{\dagger}$ & 0.659$^*$ & 0.818$^{\dagger}$ & 0.792$^{\dagger}$ & 0.661$^{\dagger}$ & 0.792$^*$ & 0.666$^{\dagger}$ & 0.831$^{\dagger}$ & 0.792$^*$ & 0.767$^*$ & \textbf{0.891}$^{\dagger}$ \\
\hline
& \textbf{AUROC} & 0.607$^{\dagger}$ & 0.564$^{\ddagger}$ & 0.619$^{\ddagger}$ & 0.527$^{\dagger}$ & 0.486$^{\dagger}$ & 0.455$^{\ddagger}$ & 0.455$^{\ddagger}$ & 0.609$^{\dagger}$ & 0.560$^{\dagger}$ & 0.577$^{\dagger}$ & \textbf{0.697}$^{\dagger}$ & 0.550$^{\ddagger}$ & 0.492$^{\ddagger}$ & 0.623$^{\ddagger}$ & \underline{0.662}$^{\ddagger}$ & 0.549$^{\dagger}$ \\
\textbf{BRCA} & \textbf{Accuracy} & 0.607$^{\dagger}$ & 0.574$^{\dagger}$ & 0.519$^{\dagger}$ & \underline{0.612}$^{\dagger}$ & 0.585$^{\dagger}$ & 0.519$^{\dagger}$ & 0.514$^{\dagger}$ & 0.536$^{\dagger}$ & 0.443$^{\dagger}$ & 0.574$^{\ddagger}$ & 0.607$^{\ddagger}$ & 0.579$^{\ddagger}$ & 0.562$^{\dagger}$ & 0.563$^{\ddagger}$ & 0.596$^{\ddagger}$ & \textbf{0.623}$^{\dagger}$ \\
& \textbf{F1-Score} & 0.253$^{\dagger}$ & 0.254$^*$ & 0.284$^{\dagger}$ & 0.255$^{\dagger}$ & 0.192$^{\dagger}$ & 0.219$^*$ & 0.192$^{\dagger}$ & 0.275$^{\dagger}$ & 0.226$^{\dagger}$ & \textbf{0.337}$^{\dagger}$ & 0.286$^{\dagger}$ & 0.206$^{\dagger}$ & \underline{0.304}$^{\dagger}$ & 0.256$^{\dagger}$ & 0.278$^{\dagger}$ & 0.192$^*$ \\
\hline
& \textbf{AUROC} & 0.621$^{\ddagger}$ & 0.539$^{\dagger}$ & 0.604$^{\dagger}$ & \underline{0.681}$^{\dagger}$ & 0.566$^{\dagger}$ & 0.570$^{\dagger}$ & 0.570$^{\dagger}$ & 0.681$^{\dagger}$ & 0.506$^{\dagger}$ & 0.582$^{\dagger}$ & 0.588$^{\ddagger}$ & 0.677$^*$ & \textbf{0.708}$^{\ddagger}$ & 0.599$^{\dagger}$ & 0.665 $^{\ddagger}$& 0.602$^{\dagger}$ \\
\textbf{CESC} & \textbf{Accuracy} & 0.557$^{\ddagger}$ & 0.519$^{\ddagger}$ & 0.557$^{\ddagger}$ & 0.592$^*$ & 0.502$^{\ddagger}$ & 0.557$^{\ddagger}$ & 0.503$^{\ddagger}$ & 0.557$^{\ddagger}$ & 0.557$^{\ddagger}$ & 0.557$^{\ddagger}$ & 0.557$^{\ddagger}$ & 0.557$^{\ddagger}$ & 0.555$^{\dagger}$ & 0.556$^{\ddagger}$ & \textbf{0.630}$^{\ddagger}$ & \underline{0.593}$^{\dagger}$ \\
& \textbf{F1-Score} & 0.555$^{\ddagger}$ & 0.516$^{\ddagger}$ & 0.555$^{\dagger}$ & 0.584$^*$ & 0.555$^{\ddagger}$ & 0.540$^{\ddagger}$ & 0.540$^{\dagger}$ & 0.550$^{\dagger}$ & 0.550$^{\dagger}$ & 0.540$^{\ddagger}$ & 0.555$^{\dagger}$ & 0.555$^{\dagger}$ & \textbf{0.703}$^{\ddagger}$ & 0.540$^{\ddagger}$ & \underline{0.629}$^{\ddagger}$ & 0.572$^{\ddagger}$ \\
\hline
& \textbf{AUROC} & 0.590$^{\dagger}$ & \textbf{0.629}$^{\dagger}$ & \textbf{0.610}$^{\dagger}$ & 0.587$^{\dagger}$ & 0.595$^{\dagger}$ & 0.572$^{\dagger}$ & 0.559$^{\dagger}$ & 0.609$^{\dagger}$ & 0.599$^{\dagger}$ & \underline{0.601}$^{\dagger}$ & 0.594$^{\dagger}$ & 0.576$^{\dagger}$ & 0.596$^{\dagger}$ & 0.564$^*$ & \textbf{0.610}$^*$ & 0.571$^{\dagger}$ \\
\textbf{NSCLC} & \textbf{Accuracy} & 0.468$^*$ & 0.482$^{\dagger}$ & 0.509$^{\dagger}$ & 0.495$^{\dagger}$ & 0.532$^{\dagger}$ & 0.486$^{\dagger}$ & 0.518$^{\dagger}$ & \underline{0.546}$^{\dagger}$ & 0.491$^*$ & 0.500$^{\dagger}$ & 0.509$^{\dagger}$ & 0.468$^{\dagger}$ & 0.502$^{\ddagger}$ & 0.307$^*$ & 0.239$^*$ & \textbf{0.781}$^{\dagger}$ \\
& \textbf{F1-Score} & 0.305$^{\dagger}$ & 0.305$^*$ & 0.313$^{\dagger}$ & 0.282$^{\dagger}$ & 0.277$^{\dagger}$ & 0.275$^*$ & 0.222$^{\dagger}$ & \underline{0.314}$^*$ & 0.041$^{\dagger}$ & 0.299$^{\dagger}$ & 0.286$^{\dagger}$ & 0.287$^{\dagger}$ & 0.270$^{\dagger}$ & 0.170$^*$ & 0.087$^*$ & \textbf{0.439}$^{\dagger}$ \\
\hline
\multicolumn{18}{c}{\textbf{UNI2~\cite{chen2024uni}}} \\
\hline
& \textbf{AUROC} & \textbf{0.918}$^{\dagger}$ & 0.783$^{\dagger}$ & 0.779$^{\dagger}$ & 0.865$^{\dagger}$ & 0.877$^{\dagger}$ & 0.879$^{\dagger}$ & 0.879$^*$ & 0.677$^{\dagger}$ & 0.836$^*$ & 0.894$^{\dagger}$ & 0.853$^{\dagger}$ & 0.811$^*$ & 0.767$^{\dagger}$ & 0.888$^*$ & 0.451$^*$ & \underline{0.897}$^*$ \\
\textbf{CAM16} & \textbf{Accuracy} & \underline{0.868}$^{\dagger}$ & 0.783$^{\dagger}$ & 0.775$^{\dagger}$ & 0.806$^{\ddagger}$ & \underline{0.868}$^{\dagger}$ & \underline{0.868}$^{\ddagger}$ & 0.865$^{\dagger}$ & 0.736$^{\dagger}$ & 0.806$^{\ddagger}$ & 0.799$^{\ddagger}$ & 0.798$^{\dagger}$ & 0.822$^{\dagger}$ & 0.752$^{\dagger}$ & 0.806$^*$ & 0.621$^*$ & \textbf{0.896}$^*$ \\
& \textbf{F1-Score} & \textbf{0.857}$^{\dagger}$ & 0.730$^{\dagger}$ & 0.736$^{\dagger}$ & 0.765$^*$ & 0.849$^{\dagger}$ & \underline{0.850}$^{\ddagger}$ & 0.849$^{\dagger}$ & 0.688$^{\dagger}$ & 0.792$^{\ddagger}$ & 0.766$^{\dagger}$ & 0.768$^{\dagger}$ & 0.793$^*$ & 0.710$^{\dagger}$ & 0.788$^*$ & 0.506$^*$ & 0.786$^{\dagger}$ \\
\hline
& \textbf{AUROC} & \underline{0.854}$^{\ddagger}$ & 0.799$^{\dagger}$ & 0.772$^{\dagger}$ & 0.848$^{\ddagger}$ & 0.852$^{\dagger}$ & \underline{0.854}$^{\ddagger}$ & 0.805$^{\ddagger}$ & 0.816$^{\dagger}$ & 0.621$^*$ & 0.694$^*$ & 0.831$^{\dagger}$ & 0.812$^{\dagger}$ & 0.627$^*$ & 0.843$^{\ddagger}$ & 0.636$^*$ & \textbf{0.856}$^{\ddagger}$ \\
\textbf{CAM17} & \textbf{Accuracy} & 0.850$^{\ddagger}$ & 0.810$^{\ddagger}$ & 0.810$^{\ddagger}$ & \underline{0.860}$^{\ddagger}$ & 0.810$^{\ddagger}$ & 0.810$^{\ddagger}$ & 0.810$^{\ddagger}$ & 0.730$^{\dagger}$ & 0.750$^*$ & 0.630$^{\dagger}$ & 0.800$^{\dagger}$ & 0.750$^*$ & 0.600$^{\dagger}$ & 0.820$^*$ & 0.700$^*$ & \textbf{0.890}$^{\ddagger}$ \\
& \textbf{F1-Score} & 0.596$^{\ddagger}$ & 0.568$^{\ddagger}$ & 0.554$^{\ddagger}$ & 0.597$^{\ddagger}$ & \underline{0.621}$^{\ddagger}$ & 0.566$^{\ddagger}$ & 0.580$^{\ddagger}$ & 0.546$^{\ddagger}$ & 0.404$^*$ & 0.354$^*$ & 0.499$^{\dagger}$ & 0.385$^{\dagger}$ & 0.325$^{\dagger}$ & 0.537$^*$ & 0.360$^*$ & \textbf{0.657}$^{\ddagger}$ \\
\hline
& \textbf{AUROC} & 0.949$^{\dagger}$ & 0.950$^{\dagger}$ & 0.954$^{\dagger}$ & 0.947$^{\dagger}$ & 0.949$^*$ & 0.945$^{\dagger}$ & 0.953$^*$ & 0.941$^{\dagger}$ & 0.940$^{\ddagger}$ & \underline{0.956}$^{\dagger}$ & 0.938$^*$ & 0.927$^{\dagger}$ & 0.936$^{\dagger}$ & 0.947$^{\ddagger}$ & 0.955$^{\ddagger}$ & \textbf{0.966}$^{\dagger}$ \\
\textbf{PANDA} & \textbf{Accuracy} & 0.774$^{\dagger}$ & 0.770$^{\dagger}$ & \underline{0.790}$^{\dagger}$ & 0.766$^{\dagger}$ & 0.765$^{\dagger}$ & 0.760$^*$ & 0.766$^{\ddagger}$ & 0.725$^*$ & 0.763$^{\dagger}$ & \textbf{0.798}$^{\ddagger}$ & 0.736$^{\dagger}$ & 0.725$^{\dagger}$ & 0.745$^{\dagger}$ & 0.758$^{\ddagger}$ & 0.775$^{\ddagger}$ & \underline{0.790}$^{\dagger}$ \\
& \textbf{F1-Score} & 0.721$^{\dagger}$ & 0.723$^{\dagger}$ & 0.745$^{\dagger}$ & 0.712$^{\dagger}$ & 0.706$^*$ & 0.702$^{\dagger}$ & 0.706$^*$ & 0.651$^*$ & 0.697$^*$ & \underline{0.759}$^{\dagger}$ & 0.662$^*$ & 0.662$^{\dagger}$ & 0.679$^{\dagger}$ & 0.678$^{\dagger}$ & 0.726$^{\ddagger}$ & \textbf{0.766}$^{\dagger}$ \\
\hline
& \textbf{AUROC} & \textbf{0.992}$^{\dagger}$ & 0.984$^*$ & 0.970$^{\dagger}$ & 0.984$^*$ & 0.963$^{\dagger}$ & 0.943$^*$ & 0.955$^*$ & 0.957$^{\ddagger}$ & 0.733$^{\dagger}$ & 0.872$^{\dagger}$ & 0.976$^{\dagger}$ & 0.877$^{\dagger}$ & 0.873$^{\ddagger}$ & 0.980$^{\dagger}$ & \textbf{0.992}$^{\dagger}$ & \underline{0.988}$^{\ddagger}$ \\
\textbf{BLCA} & \textbf{Accuracy} & 0.872$^{\dagger}$ & 0.872$^*$ & 0.872$^{\ddagger}$ & \textbf{0.936}$^{\dagger}$ & \underline{0.894}$^{\ddagger}$ & 0.880$^{\ddagger}$ & 0.887$^{\ddagger}$ & 0.872$^{\ddagger}$ & 0.872$^{\dagger}$ & 0.872$^{\dagger}$ & \underline{0.894}$^{\dagger}$ & 0.872$^{\dagger}$ & 0.873$^{\ddagger}$ & 0.872$^{\dagger}$ & 0.872$^{\dagger}$ & \textbf{0.936}$^{\dagger}$ \\
& \textbf{F1-Score} & 0.466$^{\dagger}$ & 0.466$^*$ & 0.466$^{\dagger}$ & \textbf{0.816}$^{\dagger}$ & \underline{0.614}$^{\dagger}$ & \underline{0.614}$^{\dagger}$ & 0.614$^{\dagger}$ & 0.466$^{\dagger}$ & 0.466$^{\dagger}$ & 0.466$^{\dagger}$ & \underline{0.614}$^{\dagger}$ & 0.466$^{\dagger}$ & 0.466$^{\dagger}$ & 0.466$^{\dagger}$ & 0.466$^{\dagger}$ & \textbf{0.816}$^{\dagger}$ \\
\hline
& \textbf{AUROC} & 0.467$^*$ & \underline{0.606}$^{\dagger}$ & 0.502$^*$ & 0.560$^*$ & 0.557$^*$ & 0.566$^{\dagger}$ & 0.560$^{\dagger}$ & 0.601$^{\dagger}$ & 0.522$^{\dagger}$ & 0.523$^{\dagger}$ & 0.475$^*$ & 0.502$^{\ddagger}$ & 0.463$^{\dagger}$ & 0.465$^{\ddagger}$ & \textbf{0.711}$^{\ddagger}$ & 0.572$^{\dagger}$ \\
\textbf{BRCA} & \textbf{Accuracy} & \textbf{0.623}$^{\dagger}$ & 0.601$^{\dagger}$ & 0.579$^{\ddagger}$ & 0.541$^{\dagger}$ & \textbf{0.623}$^{\dagger}$ & \underline{0.620}$^{\ddagger}$ & \textbf{0.623}$^{\dagger}$ & 0.563$^*$ & 0.612$^{\dagger}$ & 0.590$^{\ddagger}$ & 0.514$^*$ & 0.575$^{\ddagger}$ & 0.536$^*$ & 0.601$^{\ddagger}$ & 0.590$^{\ddagger}$ & \textbf{0.623}$^{\dagger}$ \\
& \textbf{F1-Score} & 0.304$^{\dagger}$ & 0.300$^*$ & \underline{0.351}$^{\dagger}$ & 0.289$^*$ & 0.192$^{\dagger}$ & 0.199$^*$ & 0.198$^{\dagger}$ & 0.317$^{\ddagger}$ & 0.190$^{\dagger}$ & 0.238$^*$ & 0.261$^*$ & 0.257$^*$ & 0.298$^{\dagger}$ & 0.315$^{\ddagger}$ & 0.331$^{\ddagger}$ & \textbf{0.354}$^{\ddagger}$ \\
\hline
& \textbf{AUROC} & 0.566$^{\ddagger}$ & 0.485$^{\dagger}$ & \underline{0.670}$^{\ddagger}$ & 0.577$^{\dagger}$ & 0.560$^{\dagger}$ & 0.565$^{\dagger}$ & 0.566$^*$ & 0.615$^{\ddagger}$ & 0.456$^{\dagger}$ & 0.566$^{\dagger}$ & 0.440$^*$ & 0.446$^{\dagger}$ & 0.467$^*$ & \textbf{0.681}$^{\ddagger}$ & 0.621$^{\ddagger}$ & 0.648$^{\dagger}$ \\
\textbf{CESC} & \textbf{Accuracy} & \underline{0.593}$^{\dagger}$ & 0.486$^{\dagger}$ & \textbf{0.630}$^{\dagger}$ & 0.519$^{\dagger}$ & 0.556$^{\dagger}$ & 0.560$^*$ & 0.555$^*$ & \textbf{0.630}$^{\dagger}$ & 0.482$^{\dagger}$ & 0.556$^{\dagger}$ & 0.481$^{\dagger}$ & 0.482$^*$ & 0.482$^{\dagger}$ & 0.556$^{\dagger}$ & 0.556$^{\dagger}$ & \textbf{0.630}$^{\dagger}$ \\
& \textbf{F1-Score} & 0.593$^{\dagger}$ & 0.486$^{\dagger}$ & \textbf{0.625}$^{\dagger}$ & 0.516$^*$ & 0.555$^{\ddagger}$ & 0.555$^*$ & 0.555$^{\dagger}$ & 0.630$^{\dagger}$ & 0.325$^{\dagger}$ & 0.540$^{\dagger}$ & 0.475$^{\dagger}$ & 0.477$^{\dagger}$ & 0.475$^{\ddagger}$ & 0.500$^*$ & 0.524$^*$ & \underline{0.616}$^{\dagger}$ \\
\hline
& \textbf{AUROC} & 0.587$^*$ & 0.583$^*$ & 0.599$^*$ & \textbf{0.654}$^{\dagger}$ & 0.610$^{\dagger}$ & 0.615$^{\dagger}$ & 0.617$^{\dagger}$ & 0.607$^{\dagger}$ & 0.593$^{\dagger}$ & 0.548$^*$ & \underline{0.624}$^{\dagger}$ & 0.615$^{\dagger}$ & 0.601$^{\dagger}$ & 0.539$^*$ & 0.581$^*$ & 0.615$^{\dagger}$ \\
\textbf{NSCLC} & \textbf{Accuracy} & 0.280$^*$ & 0.321$^*$ & 0.307$^*$ & 0.284$^*$ & 0.336$^*$ & 0.343$^*$ & 0.340$^*$ & \textbf{0.362}$^*$ & 0.344$^*$ & 0.294$^*$ & 0.303$^{\dagger}$ & 0.290$^{\dagger}$ & 0.303$^*$ & 0.312$^*$ & 0.312$^*$ & \underline{0.345}$^*$ \\
& \textbf{F1-Score} & 0.152$^*$ & 0.167$^*$ & \underline{0.201}$^*$ & 0.143$^*$ & 0.169$^*$ & 0.188$^*$ & 0.194$^*$ & \textbf{0.204}$^*$ & 0.144$^*$ & 0.152$^*$ & 0.152$^*$ & 0.152$^*$ & 0.164$^*$ & 0.182$^*$ & 0.153$^*$ & 0.182$^*$ \\
\hline
\multicolumn{18}{c}{\textbf{MUSK~\cite{musk}}} \\
\hline
& \textbf{AUROC} & 0.942$^{\ddagger}$ & 0.943$^{\ddagger}$ & 0.881$^{\ddagger}$ & 0.943$^{\ddagger}$ & 0.908$^{\dagger}$ & 0.877$^{\dagger}$ & 0.943$^{\dagger}$ & \underline{0.947}$^{\dagger}$ & 0.588$^*$ & 0.869$^{\dagger}$ & \textbf{0.957}$^{\dagger}$ & 0.906$^{\dagger}$ & 0.877$^{\dagger}$ & 0.945$^{\dagger}$ & 0.895$^{\dagger}$ & 0.910$^{\dagger}$ \\
\textbf{CAM16} & \textbf{Accuracy} & 0.868$^{\ddagger}$ & \underline{0.884}$^{\ddagger}$ & 0.814$^{\dagger}$ & 0.868$^{\ddagger}$ & 0.837$^{\dagger}$ & 0.841$^{\dagger}$ & 0.837$^{\dagger}$ & 0.837$^{\dagger}$ & 0.620$^{\dagger}$ & 0.814$^{\dagger}$ & 0.822$^{\dagger}$ & 0.820$^{\dagger}$ & 0.866$^{\dagger}$ & \underline{0.884}$^{\ddagger}$ & 0.767$^*$ & \textbf{0.898}$^{\dagger}$ \\
& \textbf{F1-Score} & \underline{0.864}$^{\ddagger}$ & \textbf{0.880}$^{\ddagger}$ & 0.805$^{\ddagger}$ & \underline{0.864}$^{\ddagger}$ & 0.829$^{\dagger}$ & 0.837$^{\dagger}$ & 0.807$^{\dagger}$ & 0.834$^{\dagger}$ & 0.383$^*$ & 0.804$^{\ddagger}$ & 0.820$^{\dagger}$ & 0.854$^{\dagger}$ & 0.839$^{\dagger}$ & \textbf{0.880}$^{\ddagger}$ & 0.764$^*$ & 0.859$^{\dagger}$ \\
\hline
& \textbf{AUROC} & 0.906$^{\ddagger}$ & 0.855$^{\ddagger}$ & 0.859$^{\dagger}$ & 0.907$^{\ddagger}$ & 0.847$^*$ & 0.847$^*$ & 0.855$^{\ddagger}$ & 0.850$^{\ddagger}$ & 0.648$^*$ & 0.830$^{\dagger}$ & 0.871$^{\ddagger}$ & 0.855$^{\ddagger}$ & 0.643$^*$ & \underline{0.911}$^{\ddagger}$ & 0.862$^*$ & \textbf{0.914}$^{\ddagger}$ \\
\textbf{CAM17} & \textbf{Accuracy} & \textbf{0.860}$^{\ddagger}$ & 0.790$^{\dagger}$ & 0.720$^*$ & 0.840$^{\ddagger}$ & 0.840$^{\dagger}$ & 0.840$^{\dagger}$ & 0.840$^{\ddagger}$ & 0.740$^*$ & 0.660$^*$ & \underline{0.850}$^{\dagger}$ & 0.792$^{\ddagger}$ & 0.780$^{\ddagger}$ & 0.417$^*$ & 0.820$^{\dagger}$ & 0.750$^*$ & 0.844$^{\ddagger}$ \\
& \textbf{F1-Score} & \underline{0.603}$^{\dagger}$ & 0.539$^*$ & 0.509$^*$ & 0.563$^{\dagger}$ & 0.575$^{\dagger}$ & 0.583$^{\dagger}$ & \underline{0.603}$^{\dagger}$ & 0.504$^*$ & 0.200$^{\dagger}$ & 0.592$^{\dagger}$ & 0.579$^{\ddagger}$ & 0.574$^{\ddagger}$ & 0.156$^*$ & 0.539$^*$ & 0.448$^*$ & \textbf{0.615}$^{\dagger}$ \\
\hline
& \textbf{AUROC} & 0.930$^{\ddagger}$ & 0.929$^{\ddagger}$ & 0.934$^{\ddagger}$ & 0.928$^{\ddagger}$ & \underline{0.935}$^{\ddagger}$ & 0.930$^{\ddagger}$ & \underline{0.935}$^{\dagger}$ & 0.922$^{\ddagger}$ & 0.920$^{\ddagger}$ & 0.934$^{\dagger}$ & 0.923$^{\ddagger}$ & 0.913$^{\ddagger}$ & 0.910$^{\ddagger}$ & 0.930$^{\ddagger}$ & 0.924$^{\ddagger}$ & \textbf{0.936}$^{\ddagger}$ \\
\textbf{PANDA} & \textbf{Accuracy} & 0.721$^{\ddagger}$ & 0.720$^{\ddagger}$ & \textbf{0.728}$^{\ddagger}$ & 0.715$^{\ddagger}$ & \underline{0.726}$^{\ddagger}$ & 0.722$^{\ddagger}$ & \textbf{0.728}$^{\ddagger}$ & 0.692$^{\ddagger}$ & 0.706$^{\ddagger}$ & 0.720$^{\ddagger}$ & 0.702$^{\ddagger}$ & 0.689$^{\ddagger}$ & 0.670$^{\ddagger}$ & 0.712$^{\ddagger}$ & 0.696$^{\ddagger}$ & \textbf{0.728}$^{\ddagger}$ \\
& \textbf{F1-Score} & 0.661$^{\ddagger}$ & 0.654$^{\ddagger}$ & \underline{0.666}$^{\ddagger}$ & 0.641$^{\ddagger}$ & \textbf{0.667}$^{\ddagger}$ & 0.657$^{\ddagger}$ & \underline{0.666}$^{\ddagger}$ & 0.636$^{\ddagger}$ & 0.627$^{\ddagger}$ & 0.653$^{\ddagger}$ & 0.631$^{\ddagger}$ & 0.629$^{\ddagger}$ & 0.607$^{\ddagger}$ & 0.647$^{\ddagger}$ & 0.643$^{\ddagger}$ & 0.657$^{\ddagger}$ \\
\hline
& \textbf{AUROC} & 0.994$^{\ddagger}$ & 0.996$^{\ddagger}$ & 0.994$^{\ddagger}$ & 0.994$^{\ddagger}$ & 0.992$^{\ddagger}$ & 0.994$^{\dagger}$ & 0.996$^{\ddagger}$ & 0.996$^{\ddagger}$ & 0.937$^{\ddagger}$ & \underline{0.998}$^{\ddagger}$ & 0.994$^{\ddagger}$ & 0.992$^{\ddagger}$ & 0.907$^{\ddagger}$ & 0.996$^{\ddagger}$ & 0.994$^{\ddagger}$ & \textbf{0.999}$^{\ddagger}$ \\
\textbf{BLCA} & \textbf{Accuracy} & 0.955$^{\ddagger}$ & 0.932$^{\ddagger}$ & 0.955$^{\ddagger}$ & 0.955$^{\ddagger}$ & 0.955$^{\ddagger}$ & 0.955$^{\ddagger}$ & 0.955$^{\ddagger}$ & 0.955$^{\ddagger}$ & 0.955$^{\ddagger}$ & \underline{0.988}$^{\ddagger}$ & 0.955$^{\ddagger}$ & 0.955$^{\ddagger}$ & 0.955$^{\ddagger}$ & 0.955$^{\ddagger}$ & 0.966$^{\ddagger}$ & \textbf{0.990}$^{\ddagger}$ \\
& \textbf{F1-Score} & 0.845$^{\ddagger}$ & 0.732$^{\dagger}$ & 0.863$^{\dagger}$ & 0.845$^{\ddagger}$ & 0.863$^{\dagger}$ & 0.827$^*$ & 0.863$^{\dagger}$ & 0.863$^{\dagger}$ & 0.845$^{\ddagger}$ & \underline{0.970}$^{\dagger}$ & 0.863$^{\dagger}$ & 0.863$^{\dagger}$ & 0.863$^{\dagger}$ & 0.845$^{\ddagger}$ & 0.903$^{\ddagger}$ & \textbf{0.973}$^{\dagger}$ \\
\hline
& \textbf{AUROC} & \underline{0.659}$^{\dagger}$ & 0.574$^*$ & 0.480$^*$ & \textbf{0.707}$^{\dagger}$ & 0.576$^{\dagger}$ & 0.574$^{\dagger}$ & 0.576$^{\dagger}$ & 0.465$^{\dagger}$ & 0.516$^{\dagger}$ & 0.440$^{\dagger}$ & 0.486$^{\dagger}$ & 0.510$^*$ & 0.372$^{\dagger}$ & 0.583$^{\dagger}$ & 0.475$^*$ & 0.586$^{\dagger}$ \\
\textbf{BRCA} & \textbf{Accuracy} & 0.596$^{\dagger}$ & 0.590$^*$ & 0.596$^{\dagger}$ & 0.601$^{\dagger}$ & 0.590$^*$ & 0.590$^*$ & 0.595$^{\dagger}$ & 0.607$^{\dagger}$ & 0.601$^{\dagger}$ & 0.546$^{\dagger}$ & 0.574$^{\dagger}$ & 0.546$^{\dagger}$ & 0.536$^{\dagger}$ & 0.585$^{\dagger}$ & \underline{0.618}$^{\dagger}$ & \textbf{0.623}$^{\dagger}$ \\
& \textbf{F1-Score} & 0.187$^{\dagger}$ & 0.199$^{\dagger}$ & 0.189$^{\dagger}$ & 0.189$^{\dagger}$ & \underline{0.270}$^{\dagger}$ & 0.189$^*$ & 0.199$^{\dagger}$ & 0.247$^{\dagger}$ & 0.211$^{\dagger}$ & 0.255$^{\dagger}$ & \textbf{0.310}$^{\dagger}$ & 0.211$^{\dagger}$ & 0.247$^{\dagger}$ & 0.199$^{\dagger}$ & 0.227$^{\dagger}$ & 0.206$^{\dagger}$ \\
\hline
& \textbf{AUROC} & 0.588$^{\dagger}$ & 0.604$^{\dagger}$ & 0.522$^*$ & 0.615$^{\dagger}$ & 0.560$^{\dagger}$ & 0.564$^{\dagger}$ & 0.588$^{\dagger}$ & \underline{0.725}$^{\ddagger}$ & 0.478$^{\dagger}$ & 0.721$^{\ddagger}$ & 0.588$^{\dagger}$ & 0.560$^{\ddagger}$ & 0.473$^*$ & 0.610$^{\dagger}$ & \underline{0.725}$^{\ddagger}$ & \textbf{0.727}$^{\dagger}$ \\
\textbf{CESC} & \textbf{Accuracy} & 0.556$^{\dagger}$ & 0.593$^{\dagger}$ & 0.556$^{\dagger}$ & \underline{0.630}$^{\ddagger}$ & 0.556$^{\ddagger}$ & 0.556$^{\dagger}$ & 0.573$^{\ddagger}$ & 0.630$^{\dagger}$ & 0.482$^*$ & \underline{0.630}$^{\dagger}$ & 0.593$^{\dagger}$ & 0.556$^{\dagger}$ & 0.407$^{\dagger}$ & 0.482$^*$ & 0.630$^{\ddagger}$ & \textbf{0.643}$^{\ddagger}$ \\
& \textbf{F1-Score} & 0.555$^{\dagger}$ & 0.590$^{\dagger}$ & 0.555$^{\dagger}$ & \underline{0.629}$^{\ddagger}$ & 0.550$^*$ & 0.555$^{\dagger}$ & 0.555$^{\dagger}$ & 0.616$^{\dagger}$ & 0.325$^*$ & \underline{0.629}$^{\ddagger}$ & 0.571$^{\dagger}$ & 0.555$^{\dagger}$ & 0.400$^{\dagger}$ & 0.481$^*$ & \textbf{0.630}$^{\ddagger}$ & 0.600$^{\ddagger}$ \\
\hline
& \textbf{AUROC} & 0.559$^{\ddagger}$ & 0.575$^{\ddagger}$ & 0.560$^{\ddagger}$ & 0.522$^{\ddagger}$ & 0.577$^{\dagger}$ & 0.580$^{\ddagger}$ & \underline{0.582}$^{\ddagger}$ & 0.578$^{\dagger}$ & 0.567$^{\ddagger}$ & 0.494$^{\ddagger}$ & 0.578$^{\dagger}$ & 0.567$^{\dagger}$ & \textbf{0.587}$^{\ddagger}$ & 0.565$^*$ & 0.552$^{\ddagger}$ & \textbf{0.587}$^{\ddagger}$ \\
\textbf{NSCLC} & \textbf{Accuracy} & 0.294$^{\ddagger}$ & 0.326$^{\dagger}$ & 0.294$^{\ddagger}$ & \textbf{0.362}$^{\ddagger}$ & 0.262$^{\dagger}$ & 0.268$^{\dagger}$ & 0.275$^{\dagger}$ & \underline{0.339}$^{\ddagger}$ & 0.298$^{\ddagger}$ & 0.262$^{\dagger}$ & 0.303$^{\dagger}$ & 0.298$^{\dagger}$ & 0.289$^{\ddagger}$ & 0.317$^*$ & 0.280$^{\ddagger}$ & 0.333$^{\ddagger}$ \\
& \textbf{F1-Score} & 0.143$^*$ & 0.168$^{\ddagger}$ & 0.176$^*$ & 0.178$^{\dagger}$ & 0.113$^*$ & 0.138$^{\dagger}$ & 0.143$^*$ & 0.166$^{\ddagger}$ & 0.129$^{\dagger}$ & 0.124$^{\dagger}$ & 0.181$^{\dagger}$ & 0.167$^{\ddagger}$ & 0.144$^*$ & \underline{0.185}$^*$ & 0.121$^{\dagger}$ & \textbf{0.193}$^{\dagger}$ \\
\hline
\end{tabular}}
\end{table*}

\begin{figure*}[t]
    \centering    
    \includegraphics[width=\textwidth]{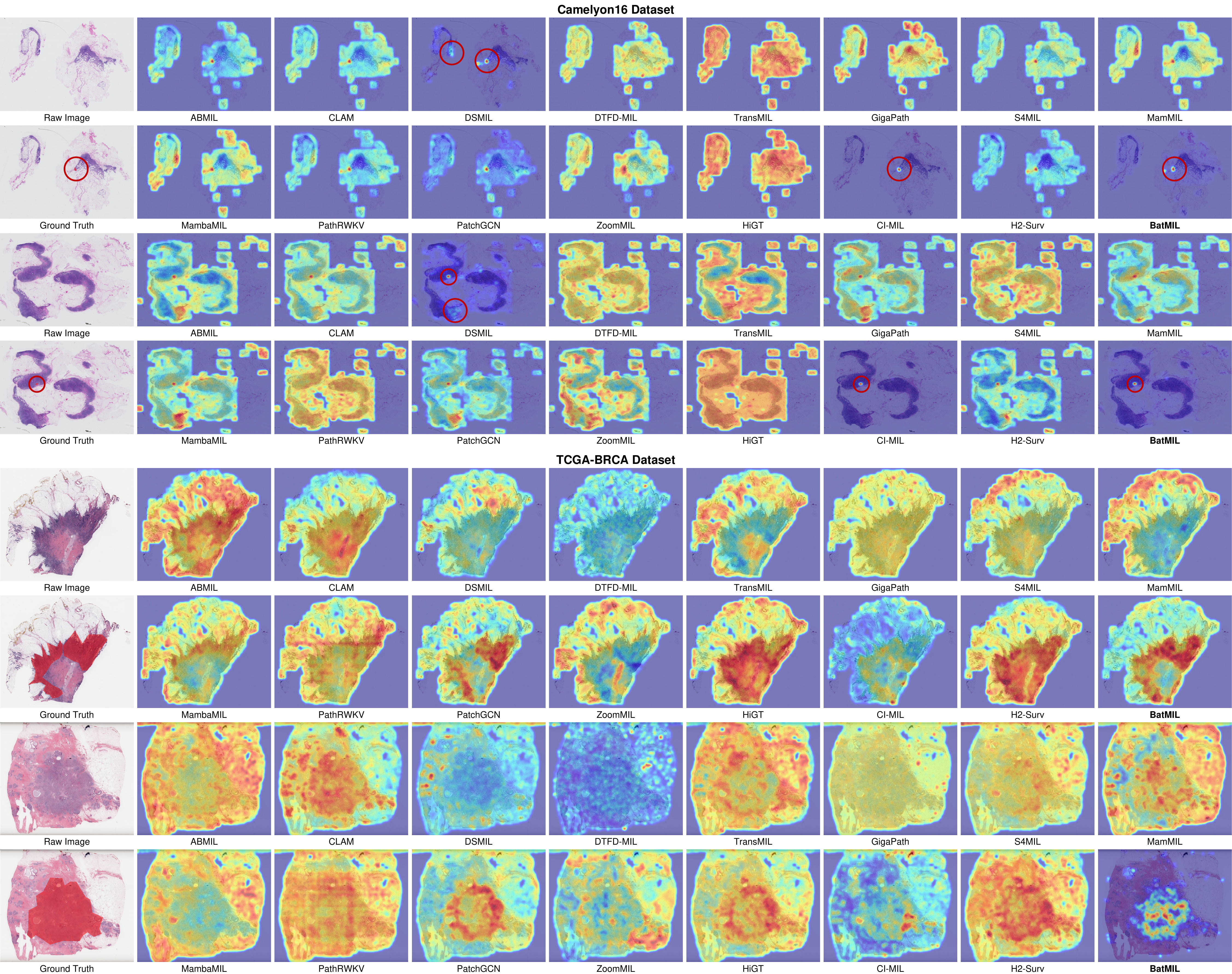}
    \caption{Heatmap visualization for different MIL methods on CAM16 and BRCA. Euclidean space-based methods exhibit highly scattered heatmaps, incorrectly attending to large swaths of benign background stroma. Some models manage partial localization but still suffer from prominent false-positive activations or dispersed background noise. BatMIL accurately identifies the multi-scale spatial relationships among tiles.}
    \label{fig:heatmap}
    
\end{figure*}

\begin{figure}[t]
    \centering
    \includegraphics[width=\linewidth]{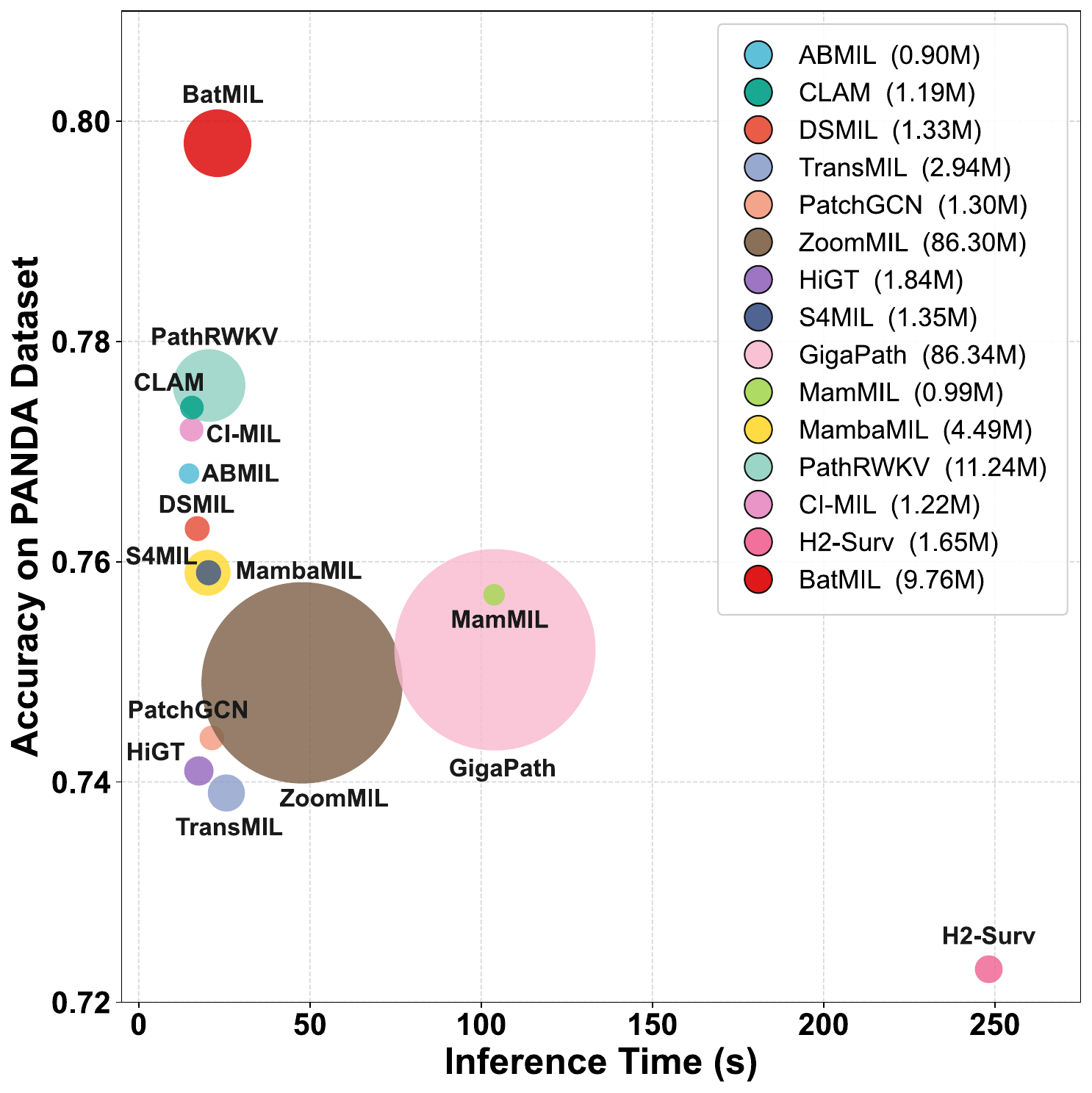}
    \caption{Comparison with SOTA methods in inference time and parameters on PANDA. The scatter plot visualizes the trade-off between diagnostic accuracy (y-axis) and computational efficiency (x-axis), with bubble size representing the total number of model parameters (in millions). BatMIL (top-left) achieves the SOTA accuracy on the PANDA dataset while maintaining a moderate parameter count (9.76M) and competitive inference speed.}
    \label{fig:efficiency}
    
\end{figure}

\subsection{Implementation Details}
We adopted three representative pathology foundation models as feature extractors: \textbf{GigaPath}~\cite{gigapath}, \textbf{UNI2}~\cite{chen2024uni}, and \textbf{MUSK}~\cite{musk}. All feature extractors were kept frozen, and all competing methods used the same extracted patch features under each encoder setting. All Euclidean parameters were optimized using AdamW~\cite{loshchilov2017decoupled} with a half-cycle cosine annealing schedule decaying to 10\% of the initial learning rate. For BatMIL, hyperbolic-space parameters were instead optimized using RiemannianAdam from the geoopt library~\cite{geoopt}, ensuring updates along the geodesics of the Poincar\'e ball. We employed validation-loss-based early stopping with a patience of 10 epochs and a maximum of 100 epochs. We conducted a hyperparameter search for the initial learning rate across $1\times10^{-4}$, $1\times10^{-5}$, and $1\times10^{-6}$. During training, we generally employed a batch size of 8 and a maximum sequence length of 4,096 tile features. For the PANDA dataset, these settings were adjusted to a batch size of 64 and a maximum of 512 tile features to accommodate dataset-specific distribution variances. Model performance was evaluated using five-fold cross-validation. Within each fold, learning-rate selection and early stopping were performed using only the validation set; the held-out test fold was not accessed during training or model selection. All experiments were implemented in Python 3.12.12 using PyTorch 2.9.0 and CUDA 12.8, running on four NVIDIA RTX 4090 GPUs. The foundational preprocessing and training pipeline was supported by UnPuzzle~\cite{UnPuzzle}.

\subsection{Comparison with SOTA Methods}
We rigorously evaluate BatMIL against fifteen SOTA methods grouped into four mainstream architectures. Specifically, \textbf{ABMIL}~\cite{abmil}, \textbf{CLAM}~\cite{clam}, \textbf{DSMIL}~\cite{dsmil}, and \textbf{DTFD-MIL}~\cite{dtfd} are attention-based methods; \textbf{PatchGCN}~\cite{patchgcn}, \textbf{ZoomMIL}~\cite{zoommil}, and \textbf{HiGT}~\cite{higt} are hierarchical methods; \textbf{TransMIL}~\cite{transmil} and \textbf{GigaPath}~\cite{gigapath} are Transformer-based methods; and \textbf{S4MIL}~\cite{s4mil}, \textbf{MambaMIL}~\cite{mambamil}, \textbf{MamMIL}~\cite{mammil}, and \textbf{PathRWKV}~\cite{chen2025pathrwkv} are SSM-based methods; \textbf{CI-MIL}~\cite{cimil} and \textbf{H2-Surv}~\cite{H2-Surv} are representation learning methods.

As shown in Tab.~\ref{tab:comparison}, BatMIL demonstrates robust performance against attention-based, hierarchical, Transformer-based, SSM-based, and representation-based baselines across GigaPath, UNI2, and MUSK features. Its clearest advantages appear on CAMELYON17, PANDA, and TCGA-BLCA, where it ranks first or near first on most encoder--metric combinations; with GigaPath features, for example, BatMIL exceeds the second-best result by 5.0 percentage points in AUROC on CAMELYON17 and by 4.6 percentage points in F1-Score on TCGA-BLCA. BatMIL also outperforms both CI-MIL and H2-Surv on all three metrics for CAMELYON17 and PANDA across all three feature extractors, indicating that its gains extend beyond existing causal and hyperbolic representation approaches. On CAMELYON16, TCGA-BRCA, TCGA-CESC, and TCGA-NSCLC, the leading method varies across metrics and feature extractors; nevertheless, BatMIL frequently achieves the highest or tied-highest Accuracy, demonstrating strong cross-encoder robustness.

\subsection{Efficiency Analysis}
Fig.~\ref{fig:efficiency} presents the performance and efficiency comparison on the PANDA dataset. The x-axis represents the inference time (s), the y-axis denotes the accuracy on the PANDA dataset, and the bubble size corresponds to the number of model parameters. BatMIL achieves the highest accuracy (79.8\%) among all evaluated methods, while maintaining a moderate parameter count of 9.76M and a competitive inference time of 23.03s. Compared to heavier architectures such as GigaPath (86.34M) and ZoomMIL (86.30M), BatMIL delivers superior accuracy with significantly lower computational overhead. Conversely, when compared to lightweight models like ABMIL (0.90M), MamMIL (0.99M), CLAM (1.19M), and S4MIL (1.35M), BatMIL yields substantial accuracy improvements with a moderate increase in inference time. Overall, BatMIL achieves an favorable trade-off between diagnostic accuracy, computational efficiency, and model complexity.

\subsection{Visualization Analysis}
To assess BatMIL's ability to localize pathological regions during WSI representation learning, we conduct Grad-CAM~\cite{gradcam}
visualization on both CAMELYON16 and TCGA-BRCA. As shown in Fig.~\ref{fig:heatmap}, we present two challenging cases from each dataset, including the raw WSI, ground-truth (GT) annotation, and Grad-CAM maps from 15 baseline methods and BatMIL. Red annotations in the GT panels indicate the pathological regions. To ensure a fair cross-model comparison, all heatmaps are generated from the final attention or aggregation layer of the corresponding model. The diffuse responses observed for several baselines are not caused by suboptimal hyperparameter tuning: the displayed cases were selected because they contain morphologically confounding regions.

On CAMELYON16, many conventional MIL methods respond broadly to benign tissue that is morphologically similar to metastatic regions. ABMIL and CLAM produce diffuse activations extending across large tissue areas, while PatchGCN partially localizes the lesions but retains non-target responses. DSMIL produces comparatively focused heatmaps because its critical-instance-guided aggregation first selects the highest-scoring instance as an anchor and then computes similarity-based attention over the remaining instances. Nevertheless, additional hotspots outside the annotated lesions remain. BatMIL likewise emphasizes informative patches, but employs feature-dependent sparse MoE routing rather than critical-instance anchoring. In both displayed cases, its strongest responses align with the annotated metastatic foci, with substantially less residual activation in surrounding benign tissue.

To evaluate localization in a more challenging diagnostic setting, we further visualize the 4-class IHC-HER2 status prediction task on TCGA-BRCA. Several baselines exhibit tissue-wide or fragmented activations; PatchGCN and HiGT partially localize the annotated tumor regions, whereas CI-MIL and H2-Surv show less consistent localization across the two cases. In contrast, BatMIL concentrates its strongest responses within the annotated tumor regions while suppressing surrounding tissue, particularly in the second case. These results provide qualitative evidence that BatMIL's localization capability extends beyond binary metastasis detection to the more challenging multi-class IHC-HER2 prediction task.

\begin{figure*}[htbp]
    
    \centering
    \includegraphics[width=\linewidth]{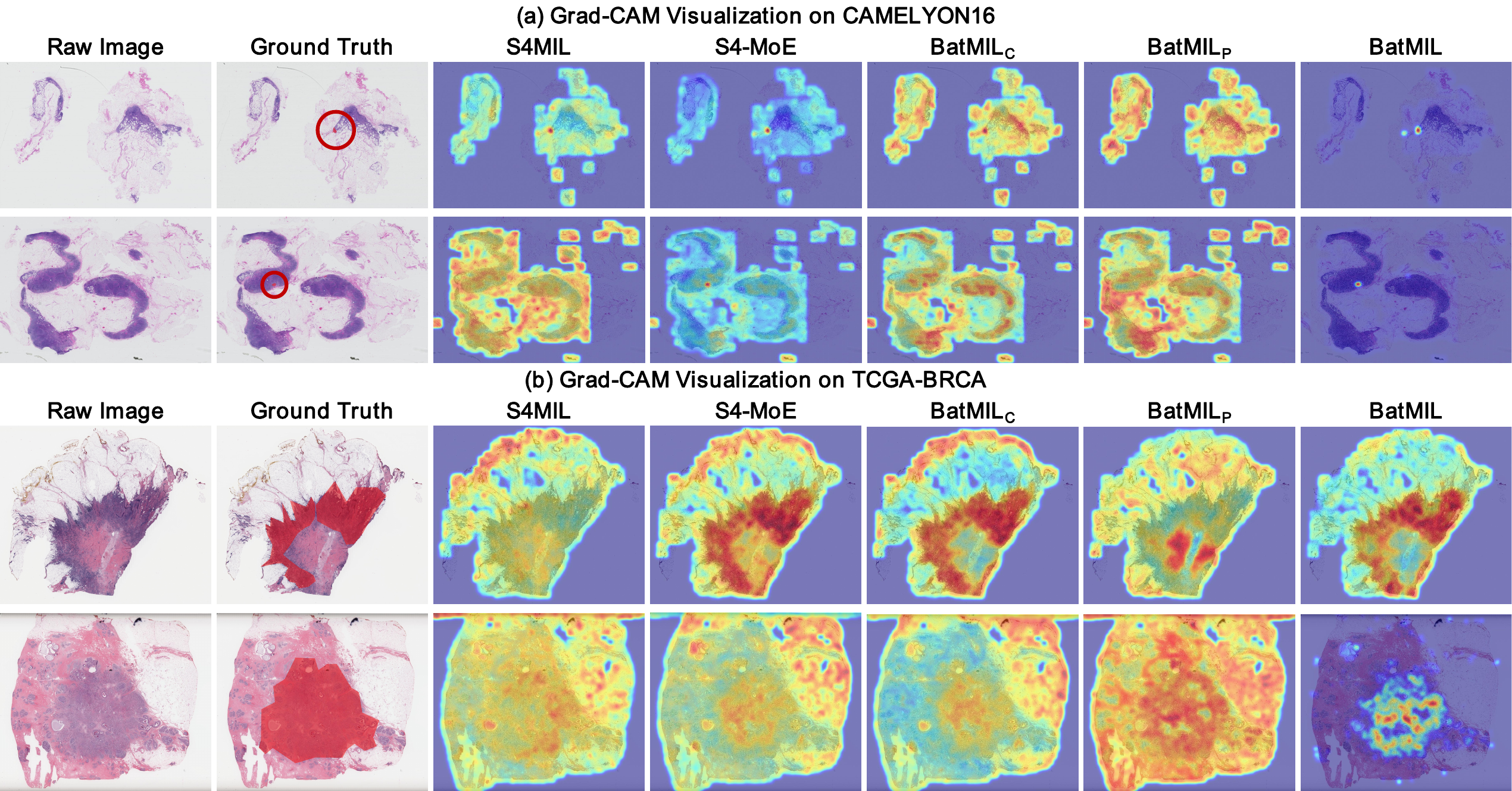}
    \caption{Heatmap visualization of ablation studies. S4MIL often disperses attention across benign tissues, while S4-MoE effectively shifts the focus toward pathological abnormalities. $\text{BatMIL}_C$ and $\text{BatMIL}_P$ exhibit scattered and redundant activations due to geometric feature interference. In contrast, BatMIL achieves the most precise attention alignment with the ground truth, successfully filtering out irrelevant background noise.}
    \label{fig:ablation}
    
\end{figure*}

%% file: Sections/5_Discussion.tex
\section{Discussion}
\subsection{Ablations on the Proposed Components}
To systematically evaluate the contribution of each architectural innovation, we conducted ablation studies (Tab.~\ref{tab:ablation1}) alongside heatmap visualization experiments (Fig.~\ref{fig:ablation}). Using S4MIL as the baseline, we first analyze the impact of the S4-MoE backbone. Quantitatively, S4-MoE demonstrates superior capability in handling complex pathological tissues, improving the AUROC by 0.053 on TCGA-BLCA and achieving an accuracy of 0.656 on TCGA-NSCLC. These improvements indicate that the MoE module effectively disentangles regional heterogeneity, preventing the parameter dilution often seen in unified networks when processing morphologically divergent tumor microenvironments. This is directly mirrored in the qualitative visualizations: while the baseline S4MIL often disperses attention across large swaths of benign tissue, the dynamic routing in S4-MoE effectively shifts the model’s focus toward pathological abnormalities, successfully filtering out background stroma.

Building upon the S4-MoE backbone, we explicitly ablate the Geometric Hybrid Strategy (GHS) module by comparing the S4-MoE variant against the final BatMIL. We compared five strategies. $\text{BatMIL}_C$ directly concatenates the Euclidean and tangent-space hyperbolic representations; $\text{BatMIL}_P$ concatenates the two representations and applies a learnable linear projection to restore the original embedding dimensionality; $\text{BatMIL}_G$ employs an instance-dependent gating mechanism to adaptively combine them; $\text{BatMIL}_A$ uses cross-attention to model interactions between the two representations; and the final BatMIL adopts a globally shared, feature-wise weighted-addition strategy to balance their contributions.

As shown in Tab.~\ref{tab:ablation1}, all the ablation variants failed to overcome BatMIL on the overall performance. Specifically, although $\text{BatMIL}_C$ and $\text{BatMIL}_P$ retain partial benefits from the dual-branch design, their improvements are inconsistent across datasets and metrics. $\text{BatMIL}_C$ directly concatenates the Euclidean and tangent-space hyperbolic representations, which doubles the classifier input dimension from $m$ to $2m$ and may introduce feature redundancy. $\text{BatMIL}_P$ applies a projection after fusion, but this additional transformation does not yield consistent gains over S4-MoE. $\text{BatMIL}_G$ and $\text{BatMIL}_A$ exhibit larger performance fluctuations. In contrast, BatMIL consistently achieves the highest Accuracy on all seven datasets. These results indicate that weighted addition provides the most stable and effective fusion among the evaluated strategies.

Visually, Fig.~\ref{fig:ablation} further supports this observation. S4MIL and S4-MoE produce diffuse activations over large benign regions, indicating that Euclidean-only representations are insufficient to suppress confounding tissue responses. Although $\text{BatMIL}_C$ and $\text{BatMIL}_P$ incorporate hyperbolic features, their heatmaps remain scattered, suggesting that direct concatenation or post-fusion projection does not harmonize the two geometric branches effectively. The nonlinear fusion variants $\text{BatMIL}_G$ and $\text{BatMIL}_A$ also fail to provide clean localization, with substantial residual background responses and unstable emphasis on the true lesions. In contrast, BatMIL produces the cleanest and most focused heatmaps, with the strongest activations concentrated within the metastatic or malignant regions, and markedly fewer off-target responses
in surrounding normal tissue. We interpret the cleaner localization in Fig.~\ref{fig:ablation} as empirical evidence associated with more stable dual-branch fusion.

Furthermore, we evaluated the efficiency of each component in BatMIL. As shown in Tab.~\ref{tab:module_efficiency}, the S4 backbone contains only 0.56M parameters and requires 6.86 s for inference, while the MoE routing module accounts for the majority of the model capacity with 8.14M parameters and requires 14.23 s. In contrast, despite incorporating hyperbolic mappings and M\"obius transformations, the GHS module introduces only 1.06M parameters and 0.69 s of inference time. These results demonstrate that MoE routing constitutes the primary computational burden of BatMIL.

\begin{table}[t]
\caption{Ablations on the proposed modules. (Note: Standard deviations $\sigma$ are represented by symbols: $^{\ddagger}$ for $\sigma \le 0.005$, $^{\dagger}$ for $0.005 < \sigma \le 0.015$, and $^*$ for $\sigma > 0.015$)}
\label{tab:ablation1}
\centering
\setlength{\tabcolsep}{1pt}
\renewcommand{\arraystretch}{1.2}
\resizebox{\linewidth}{!}{
\begin{tabular}{c|c|cc|cccc|c}
\hline
\textbf{Dataset} & \textbf{Metric} & \textbf{S4MIL} & \textbf{S4-MoE} & \textbf{BatMIL$_C$} & \textbf{BatMIL$_P$} & \textbf{BatMIL$_G$} & \textbf{BatMIL$_A$} & \textbf{BatMIL} \\
\hline
& \textbf{AUROC} & 0.989$^*$ & 0.989$^*$ & 0.989$^{\dagger}$ & 0.989$^{\dagger}$ & 0.591$^*$ & 0.974$^{\dagger}$ & \textbf{0.996}$^{\ddagger}$ \\
\textbf{CAM16} & \textbf{Accuracy} & 0.984$^{\ddagger}$ & 0.984$^{\dagger}$ & 0.984$^{\dagger}$ & 0.978$^{\dagger}$ & 0.620$^*$ & 0.953$^{\dagger}$ & \textbf{0.988}$^{\ddagger}$ \\
& \textbf{F1-Score} & \textbf{0.983}$^{\dagger}$ & 0.977$^{\dagger}$ & 0.973$^{\dagger}$ & 0.978$^{\ddagger}$ & 0.383$^*$ & 0.949$^{\dagger}$ & 0.977$^*$ \\
\hline
& \textbf{AUROC} & 0.541$^*$ & 0.450$^{\dagger}$ & 0.547$^{\dagger}$ & 0.457$^{\dagger}$ & 0.759$^*$ & 0.711$^{\dagger}$ & \textbf{0.778}$^*$ \\
\textbf{CAM17} & \textbf{Accuracy} & 0.646$^{\dagger}$ & 0.656$^*$ & 0.705$^{\dagger}$ & 0.636$^{\dagger}$ & 0.768$^*$ & 0.737$^{\dagger}$ & \textbf{0.778}$^*$ \\
& \textbf{F1-Score} & 0.244$^*$ & 0.198$^{\dagger}$ & 0.505$^*$ & 0.426$^{\dagger}$ & 0.377$^{\dagger}$ & 0.359$^{\dagger}$ & \textbf{0.507}$^*$ \\
\hline
& \textbf{AUROC} & 0.940$^{\ddagger}$ & 0.949$^{\dagger}$ & 0.950$^{\dagger}$ & 0.941$^{\dagger}$ & 0.845$^*$ & 0.818$^*$ & \textbf{0.960}$^{\ddagger}$ \\
\textbf{PANDA} & \textbf{Accuracy} & 0.759$^{\ddagger}$ & 0.797$^{\ddagger}$ & 0.770$^{\dagger}$ & 0.774$^{\ddagger}$ & 0.517$^*$ & 0.339$^*$ & \textbf{0.798}$^{\ddagger}$ \\
& \textbf{F1-Score} & 0.707$^{\ddagger}$ & 0.724$^{\dagger}$ & 0.715$^{\ddagger}$ & 0.724$^{\ddagger}$ & 0.210$^*$ & 0.136$^*$ & \textbf{0.748}$^{\dagger}$ \\
\hline
& \textbf{AUROC} & 0.942$^{\dagger}$ & 0.995$^{\dagger}$ & \textbf{0.997}$^{\ddagger}$ & 0.996$^{\dagger}$ & 0.997$^{\ddagger}$ & 0.997$^{\ddagger}$ & \textbf{0.997}$^{\ddagger}$ \\
\textbf{BLCA} & \textbf{Accuracy} & 0.921$^{\dagger}$ & 0.931$^{\dagger}$ & 0.954$^{\dagger}$ & 0.943$^{\ddagger}$ & 0.898$^{\dagger}$ & 0.966$^{\ddagger}$ & \textbf{0.977}$^{\dagger}$ \\
& \textbf{F1-Score} & 0.661$^{\dagger}$ & 0.731$^{\dagger}$ & 0.845$^{\dagger}$ & 0.792$^{\ddagger}$ & 0.473$^{\dagger}$ & 0.802$^{\ddagger}$ & \textbf{0.891}$^{\dagger}$ \\
\hline
& \textbf{AUROC} & 0.577$^{\dagger}$ & 0.634$^{\dagger}$ & 0.601$^{\dagger}$ & \textbf{0.655}$^{\ddagger}$ & 0.480$^*$ & 0.472$^*$ & 0.549$^{\dagger}$ \\
\textbf{BRCA} & \textbf{Accuracy} & 0.574$^{\ddagger}$ & 0.584$^{\dagger}$ & 0.617$^{\ddagger}$ & 0.562$^{\dagger}$ & 0.620$^*$ & 0.620$^*$ & \textbf{0.623}$^{\dagger}$ \\
& \textbf{F1-Score} & \textbf{0.337}$^{\dagger}$ & 0.224$^{\ddagger}$ & 0.204$^{\ddagger}$ & 0.287$^{\ddagger}$ & 0.192$^*$ & 0.192$^*$ & 0.192$^*$ \\
\hline
& \textbf{AUROC} & 0.582$^{\dagger}$ & 0.596$^{\dagger}$ & 0.596$^{\dagger}$ & 0.593$^{\dagger}$ & 0.511$^*$ & 0.591$^*$ & \textbf{0.602}$^{\dagger}$ \\
\textbf{CESC} & \textbf{Accuracy} & 0.557$^{\ddagger}$ & 0.575$^{\dagger}$ & 0.567$^{\dagger}$ & 0.566$^{\dagger}$ & 0.482$^*$ & 0.556$^{\dagger}$ & \textbf{0.593}$^{\dagger}$ \\
& \textbf{F1-Score} & 0.540$^{\ddagger}$ & 0.544$^{\dagger}$ & 0.556$^{\dagger}$ & 0.550$^{\dagger}$ & 0.325$^*$ & 0.540$^{\dagger}$ & \textbf{0.572}$^{\ddagger}$ \\
\hline
& \textbf{AUROC} & 0.601$^{\dagger}$ & \textbf{0.603}$^{\dagger}$ & 0.587$^{\dagger}$ & 0.501$^{\dagger}$ & 0.514$^*$ & 0.374$^*$ & 0.571$^{\dagger}$ \\
\textbf{NSCLC} & \textbf{Accuracy} & 0.500$^{\dagger}$ & 0.656$^*$ & 0.716$^{\dagger}$ & 0.662$^{\dagger}$ & 0.266$^*$ & 0.266$^*$ & \textbf{0.781}$^{\dagger}$ \\
& \textbf{F1-Score} & 0.299$^{\dagger}$ & 0.440$^{\dagger}$ & 0.425$^{\dagger}$ & 0.420$^*$ & 0.047$^*$ & 0.085$^*$ & \textbf{0.439}$^{\dagger}$ \\
\hline
\end{tabular}}
\end{table}

\begin{table}[t]
\caption{Per-module parameter count and inference time of BatMIL on the PANDA dataset.}
\label{tab:module_efficiency}
\centering
\renewcommand{\arraystretch}{1.2}
\resizebox{\linewidth}{!}{
\begin{tabular}{c|c|c}
\hline
\textbf{Module} & \textbf{Params (M)} & \textbf{Inference Time (s)} \\
\hline
S4 Backbone & 0.56 & 6.86 \\
MoE Routing & 8.14 & 14.23 \\
GHS Module & 1.06 & 0.69 \\
\hline
\end{tabular}}

\end{table}

\subsection{Hyperbolic Distance Embedding Analysis}
The manifold curvature $hyp_c$ is the defining parameter of the Poincaré ball model, directly controlling its capacity to embed tree-like structures. To study its effect on BatMIL, we evaluated four fixed curvature values: $hyp_c \in \{0.05, 0.075, 0.1, 0.2\}$, and a learnable curvature variant $hyp_{dyn}$ that is optimized during training.

As demonstrated in Tab.~\ref{tab:ablation2}, BatMIL achieves optimal performance across most datasets when $hyp_c = 0.1$. For example, on the CAMELYON16 dataset, the AUROC peaks at 0.996 for $hyp_c = 0.1$, outperforming the configurations of 0.05, 0.075, and 0.2 by margins of 0.020, 0.011, and 0.025, respectively. Similar performance peaks are observed in the PANDA, TCGA-BLCA, and TCGA-NSCLC datasets for this curvature value. This behavior highlights a fundamental geometric trade-off. When $hyp_c$ is too small (e.g., 0.05), the manifold approaches a flatter geometry, limiting its capacity to represent exponentially expanding hierarchical structures. Conversely, an excessively large curvature (e.g., 0.2) may over-distort local geometric relationships and weaken the discrimination of fine-grained morphological patterns. Our choice $hyp_c = 0.1$ strikes the best empirical trade-off among the tested values, preserving the macroscopic hierarchical topology while maintaining microscopic morphological fidelity.

We further evaluate a learnable-curvature variant, in which $hyp_c$ is jointly optimized with the network parameters. The learned curvature converges close to 0.1, ranging from 0.100105 to 0.116160 across the seven datasets. However, dynamically optimizing curvature does not yield consistent gains: compared with fixed $hyp_c=0.1$, the learnable variant improves only four dataset--metric combinations while degrading seven. These mixed results suggest that joint curvature optimization may favor individual metrics at the expense of overall predictive stability. Therefore, we adopt fixed $hyp_c=0.1$ as the default setting.

\begin{table}
\caption{Ablations on the $hyp_c$ settings. (Note: Standard deviations $\sigma$ are represented by symbols: $^{\ddagger}$ for $\sigma \le 0.005$, $^{\dagger}$ for $0.005 < \sigma \le 0.015$, and $^*$ for $\sigma > 0.015$.)}
\label{tab:ablation2}
\centering
\setlength{\tabcolsep}{1pt}
\renewcommand{\arraystretch}{1.2}
\resizebox{\linewidth}{!}{
\begin{tabular}{c|c|cccc|cc}
\hline
\textbf{Dataset} & \textbf{Metric} & \textbf{$hyp_c=0.05$} & \textbf{$hyp_c=0.075$} & \textbf{$hyp_c=0.1$} & \textbf{$hyp_c=0.2$} & \textbf{$hyp_{dyn}$} & \textbf{$BEST_{hyp_c}$} \\
\hline
& \textbf{AUROC} & 0.976$^{\ddagger}$ & 0.985$^{\ddagger}$ & \textbf{0.996}$^{\ddagger}$ & 0.971$^{\ddagger}$ & 0.996$^{\dagger}$ &  \\
\textbf{CAM16} & \textbf{Accuracy} & 0.967$^{\ddagger}$ & 0.977$^{\ddagger}$ & \textbf{0.988}$^{\ddagger}$ & 0.963$^{\ddagger}$ & 0.988$^{\ddagger}$ & 0.100105 \\
& \textbf{F1-Score} & 0.967$^{\ddagger}$ & 0.970$^{\ddagger}$ & \textbf{0.977}$^*$ & 0.973$^{\ddagger}$ & 0.977$^{\dagger}$ & \\
\hline
& \textbf{AUROC} & 0.754$^*$ & 0.759$^{\ddagger}$ & \textbf{0.778}$^*$ & 0.754$^{\dagger}$ & 0.796$^{\dagger}$ &  \\
\textbf{CAM17} & \textbf{Accuracy} & 0.755$^{\dagger}$ & 0.768$^*$ & \textbf{0.778}$^*$ & 0.770$^*$ & 0.778$^{\dagger}$ & 0.116160 \\
& \textbf{F1-Score} & 0.449$^*$ & 0.503$^{\dagger}$ & \textbf{0.507}$^*$ & 0.503$^{\dagger}$ & 0.385$^{\dagger}$ &  \\
\hline
& \textbf{AUROC} & 0.959$^{\ddagger}$ & 0.954$^{\ddagger}$ & \textbf{0.960}$^{\ddagger}$ & 0.955$^{\ddagger}$ & 0.939$^{\dagger}$ &  \\
\textbf{PANDA} & \textbf{Accuracy} & 0.786$^{\ddagger}$ & 0.789$^{\ddagger}$ & \textbf{0.798}$^{\ddagger}$ & 0.791$^{\ddagger}$ & 0.748$^{\dagger}$ & 0.104120 \\
& \textbf{F1-Score} & 0.737$^{\dagger}$ & 0.735$^{\ddagger}$ & \textbf{0.748}$^{\dagger}$ & 0.740$^{\ddagger}$ & 0.677$^{\dagger}$ &  \\
\hline
& \textbf{AUROC} & 0.976$^{\ddagger}$ & 0.983$^{\ddagger}$ & \textbf{0.997}$^{\ddagger}$ & 0.979$^{\ddagger}$ & 0.997$^{\ddagger}$ &  \\
\textbf{BLCA} & \textbf{Accuracy} & \textbf{0.977}$^{\dagger}$ & \textbf{0.977}$^{\dagger}$ & \textbf{0.977}$^{\dagger}$ & 0.966$^{\dagger}$ & 0.977$^{\dagger}$ & 0.100256 \\
& \textbf{F1-Score} & 0.877$^{\ddagger}$ & 0.890$^{\dagger}$ & \textbf{0.891}$^{\dagger}$ & 0.871$^{\dagger}$ & 0.891$^{\dagger}$ &  \\
\hline
& \textbf{AUROC} & 0.536$^{\dagger}$ & 0.545$^{\dagger}$ & \textbf{0.549}$^{\dagger}$ & 0.545$^{\dagger}$ & 0.595$^*$ &  \\
\textbf{BRCA} & \textbf{Accuracy} & 0.615$^{\dagger}$ & 0.619$^{\dagger}$ & \textbf{0.623}$^{\dagger}$ & 0.610$^{\dagger}$ & 0.623$^{\dagger}$ & 0.100961 \\
& \textbf{F1-Score} & \textbf{0.208}$^{\dagger}$ & 0.191$^{\dagger}$ & 0.192$^*$ & 0.199$^{\dagger}$ & 0.192$^*$ &  \\
\hline
& \textbf{AUROC} & 0.598$^{\dagger}$ & 0.600$^{\ddagger}$ & \textbf{0.602}$^{\dagger}$ & 0.596$^{\dagger}$ & 0.648$^{\dagger}$ &  \\
\textbf{CESC} & \textbf{Accuracy} & 0.587$^{\dagger}$ & 0.592$^{\dagger}$ & \textbf{0.593}$^{\dagger}$ & \textbf{0.593}$^{\ddagger}$ & 0.571$^{\dagger}$ & 0.103142 \\
& \textbf{F1-Score} & 0.565$^{\ddagger}$ & 0.566$^{\ddagger}$ & \textbf{0.572}$^{\ddagger}$ & 0.564$^{\ddagger}$ & 0.552$^{\dagger}$ &  \\
\hline
& \textbf{AUROC} & 0.565$^{\dagger}$ & 0.560$^{\dagger}$ & \textbf{0.571}$^{\dagger}$ & 0.569$^{\dagger}$ & 0.546$^{\dagger}$ &  \\
\textbf{NSCLC} & \textbf{Accuracy} & 0.773$^{\dagger}$ & 0.770$^{\dagger}$ & \textbf{0.781}$^{\dagger}$ & 0.778$^{\dagger}$ & 0.787$^{\dagger}$ & 0.104337 \\
& \textbf{F1-Score} & 0.427$^{\dagger}$ & 0.427$^{\dagger}$ & \textbf{0.439}$^{\dagger}$ & 0.434$^{\dagger}$ & 0.439$^{\dagger}$ &  \\
\hline
\end{tabular}}

\end{table}

\subsection{The Interpretability of MoE Routing Mechanisms}
\begin{figure}[t]
    \centering
    \includegraphics[width=\linewidth]{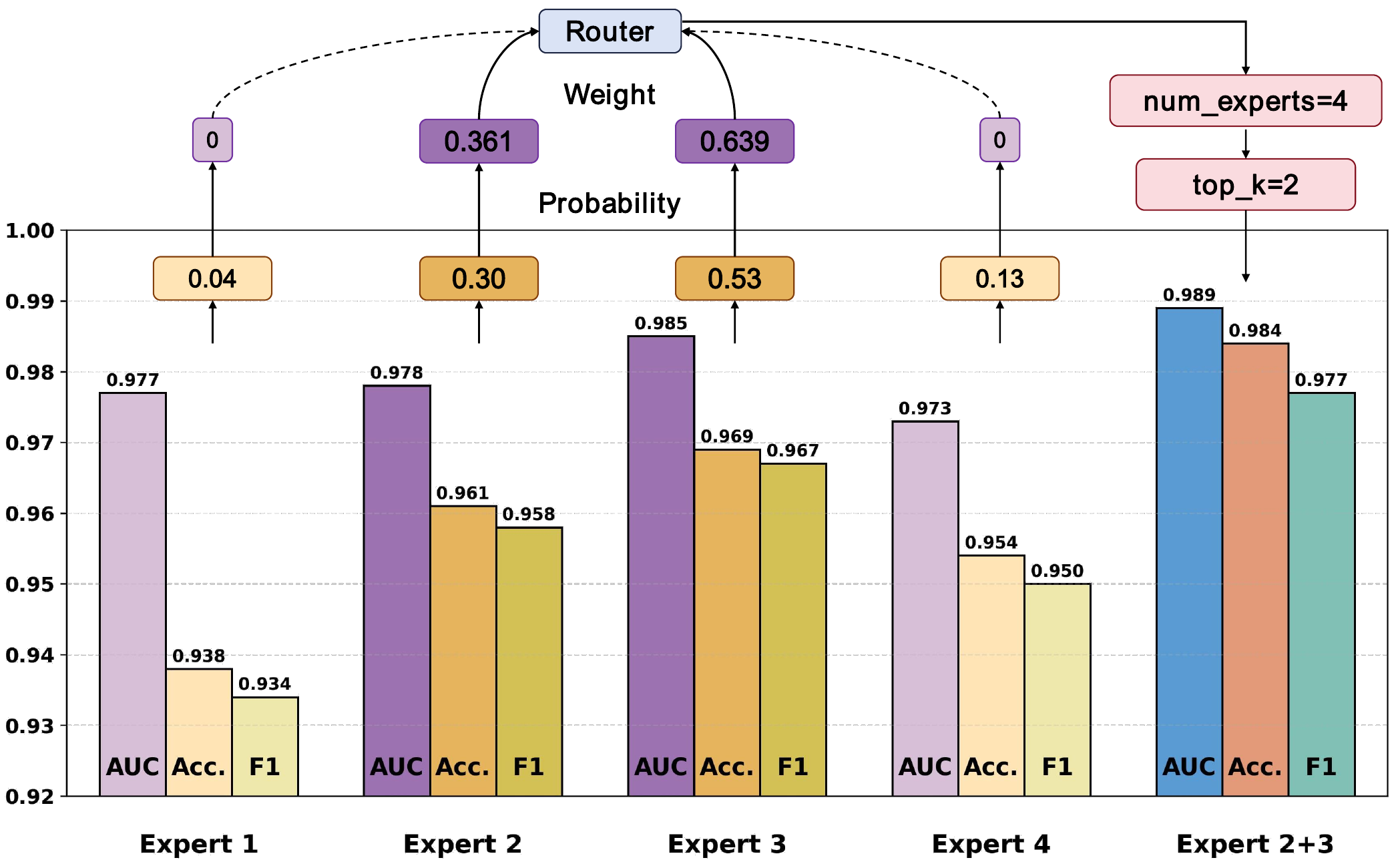}
    \caption{Interpretability of the MoE routing mechanism on the CAMELYON16 dataset. The bar chart compares the individual performance of four experts against their fused performance. The router (num-experts=4, top-k=2) adaptively assigns the highest probabilities (0.53 and 0.30) to the two best-performing experts (Expert 3 and Expert 2). After weight renormalization (0.639 and 0.361, respectively), the sparse fusion of these two experts achieves the highest overall performance, demonstrating the advantage of selective expert activation over single-expert or dense routing strategies.}
    \label{fig:expert}
\end{figure}

To examine whether the MoE router learns meaningful expert selection, we analyze both the standalone performance of each expert and the routing behavior on the CAMELYON16 dataset (Fig.~\ref{fig:expert}). When evaluated independently, the four experts show different levels of discriminative ability, with Expert 3 achieving the best performance, followed by Expert 2, Expert 4, and Expert 1. This performance hierarchy is reflected in the learned routing probabilities. For each input token, our MoE module activates only the top-2 experts among four candidates. The router assigns the highest probabilities to Expert 3 and Expert 2, with scores of 0.53 and 0.30, respectively. After top-2 selection, their routing weights are renormalized to 0.639 and 0.361, while Expert 1 and Expert 4 are suppressed. This indicates that the router does not select experts arbitrarily, but learns to select complementary experts whose combination improves representation quality beyond what any single expert can achieve. More importantly, the top-2 fusion strategy outperforms any individual expert. The fused MoE output achieves an AUROC of 0.989, an accuracy of 0.984, and an F1-score of 0.977, surpassing the best single expert, Expert 3. This suggests that Expert 2 provides complementary information that further improves the representation, whereas weaker experts may introduce noisy or less relevant features. Therefore, the sparse top-2 routing mechanism provides an interpretable means of selectively combining complementary expert representations while avoiding unnecessary expert activation.

To further investigate whether this routing behavior generalizes across diagnostic scenarios, we extend the analysis to the other six datasets in Tab.~\ref{tab:expert}. As each dataset is trained independently, expert indices (E1--E4) are permutation-invariant identifiers and do not represent semantically aligned experts across datasets. We therefore analyze cross-dataset behavior through the routing distribution. The results reveal substantial dataset-dependent variation: CAMELYON16 and CAMELYON17 exhibit relatively concentrated routing; PANDA, TCGA-BLCA, and TCGA-NSCLC show more distributed routing; and routing becomes more balanced on TCGA-BRCA and TCGA-CESC. Notably, top-$k=2$ aggregation outperforms the best individual expert on six of the seven datasets, suggesting that sparse routing generally benefits from combining complementary expert representations.

\begin{table}[t]
\caption{Routing probabilities and expert performance across datasets. We report the average top-2 routing weights, individual-expert AUROC, and routed top-$k=2$ AUROC.}
\label{tab:expert}
\centering
\setlength{\tabcolsep}{1pt}
\renewcommand{\arraystretch}{1.2}
\resizebox{\linewidth}{!}{
\begin{tabular}{c|cccc|c|c|cccc|c}
\hline
\multirow{2}{*}{\textbf{Dataset}} & \multicolumn{5}{c|}{\textbf{Routing Probability}} & \multirow{2}{*}{\textbf{Total}} & \multicolumn{5}{c}{\textbf{Individual Expert AUROC}} \\
\cline{2-6} \cline{8-12}
 & \textbf{E1} & \textbf{E2} & \textbf{E3} & \textbf{E4} & \textbf{Top 2} & & \textbf{E1} & \textbf{E2} & \textbf{E3} & \textbf{E4} & \textbf{Top-k=2} \\
\hline
\textbf{CAM16}       & 0.04 & \textbf{0.30} & \textbf{0.53} & 0.13 & E3+E2 & 0.83 & 0.977 & 0.978 & \textbf{0.985} & 0.973 & \textbf{0.989} \\
\textbf{CAM17}       & 0.05 & \textbf{0.31} & \textbf{0.52} & 0.12 & E3+E2 & 0.83 & 0.512 & 0.538 & \textbf{0.559} & 0.525 & 0.450 \\
\textbf{PANDA}            & 0.15 & \textbf{0.38} & 0.20 & \textbf{0.27} & E2+E4 & 0.65 & 0.901 & \textbf{0.935} & 0.919 & 0.928 & \textbf{0.949} \\
\textbf{BLCA}        & \textbf{0.30} & \textbf{0.38} & 0.20 & 0.12 & E1+E2 & 0.68 & 0.948 & 0.963 & \textbf{0.976} & 0.955 & \textbf{0.995} \\
\textbf{BRCA}        & \textbf{0.32} & 0.19 & 0.19 & \textbf{0.30} & E1+E4 & 0.62 & \textbf{0.548} & 0.541 & 0.530 & 0.537 & \textbf{0.634} \\
\textbf{CESC}        & \textbf{0.33} & 0.19 & 0.18 & \textbf{0.30} & E1+E4 & 0.63 & \textbf{0.570} & 0.563 & 0.555 & 0.568 & \textbf{0.596} \\
\textbf{NSCLC}       & 0.12 & 0.18 & \textbf{0.40} & \textbf{0.30} & E3+E4 & 0.70 & 0.524 & 0.538 & \textbf{0.555} & 0.530 & \textbf{0.603} \\
\hline
\end{tabular}}

\end{table}

\subsection{Limitations and Future Work}
Despite establishing a geometry-embedding paradigm, BatMIL exhibits certain limitations. In tasks where diagnostic evidence is extremely sparse and highly localized, BatMIL's performance is slightly lower than classical attention models. This discrepancy implies that while BatMIL excels at modeling macro-architectural hierarchy and global context, its robust global prior may over-smooth highly localized cellular signals. Moreover, the MoE routing module constitutes the major computational burden of BatMIL. Improving expert-routing efficiency through more lightweight architectures or more efficient mechanisms is important for future work. In addition, learnable curvature exhibits unstable optimization, motivating more robust curvature adaptation strategies in future work. Finally, although the experiments cover diverse cancer types, diagnostic criteria, and task complexities on the classification task, broader validation on downstream tasks such as survival prediction, retrieval, visual question answering, and report generation is required to further establish the generalizability of BatMIL as a WSI representation framework.

%% file: Sections/6_Conclusion.tex
\section{Conclusion}
In this paper, we proposed BatMIL, a novel framework designed to address the limitations of existing MIL methods that embed pathological tissue representations in a single Euclidean space while overlooking hierarchical structure and regional heterogeneity. Specifically, we designed a hybrid hyperbolic-Euclidean geometric representation that jointly embeds WSI features into dual geometric spaces to facilitate hierarchical embedding. Building on this, we designed an S4-MoE module that efficiently encodes long-range dependencies with linear computational complexity, and dynamically adapts to the pronounced regional heterogeneity of pathological tissues through a MoE routing mechanism. Extensive experiments on seven WSI datasets across six cancer types demonstrate that BatMIL consistently outperforms fifteen SOTA MIL methods across six downstream tasks. Comprehensive ablation studies and interpretability analyses further confirm the effectiveness of each component and the rationality of the weighted-addition strategy. In summary, BatMIL provides an effective hybrid approach in computational pathology, and could provide a promising direction for the future development of more accurate and interpretable clinical decision-support systems.